\documentclass[lettersize,journal]{IEEEtran}
\usepackage{amsmath,amsfonts}
\usepackage{algorithmic}
\usepackage{algorithm}
\usepackage{array}
\usepackage{booktabs} 
\usepackage{multirow} 
\usepackage{xcolor,colortbl}
\usepackage{threeparttable}

\usepackage{textcomp}
\usepackage{stfloats}
\usepackage{url}
\usepackage{verbatim}
\usepackage{graphicx}
\usepackage{cite}
\usepackage{amssymb}
\usepackage{mathrsfs}
\usepackage{multirow}
\usepackage{makecell} 
\usepackage{subcaption} 
\usepackage{booktabs}
\newtheorem{definition}{Definition}
\begin{document}

\title{Guided Collaboration in Heterogeneous LLM-Based Multi-Agent Systems via Entropy-Based Understanding Assessment and Experience Retrieval}

\author{Linlin Wang, Tianqing Zhu*,~\IEEEmembership{ Member,~IEEE,} 
Laiqiao Qin, 
Longxiang Gao,~\IEEEmembership{ Senior Member,~IEEE,} 
Wanlei Zhou,~\IEEEmembership{ Life Fellow,~IEEE}

        % <-this % stops a space
\thanks{*corresponding author.  Linlin Wang, Tianqing Zhu, Laiqiao Qin and Wanlei Zhou are with the Faculty of Data Science, City University of Macau, Macao, China (e-mail:~linlinwang.cityu@gmail.com; tqzhu@cityu.edu.mo; isqlq@outlook.com; wlzhou@cityu.edu.mo)}% <-this % stops a space

\thanks{Longxiang Gao is with the Key Laboratory of Computing Power Network and Information Security, Ministry of Education, Shandong Computer Science Center, Qilu University of Technology (Shandong Academy of Sciences), Jinan, China, and also with the Shandong Provincial Key Laboratory of Computing Power Internet and Service Computing, Shandong Fundamental Research Center for Computer Science, Jinan, China (e-mail: gaolx@sdas.org)}}


\markboth{Journal of \LaTeX\ Class Files,~Vol.~14, No.~8, August~2021}%
{Shell \MakeLowercase{\textit{et al.}}: A Sample Article Using IEEEtran.cls for IEEE Journals}


\maketitle

\begin{abstract}
With recent breakthroughs in large language models (LLMs) for reasoning, planning, and complex task generation, artificial intelligence systems are transitioning from isolated single-agent architectures to multi-agent systems with collaborative intelligence. However, in heterogeneous multi-agent systems (HMAS), capability differences among agents gives rise to a consistent cognitive problems, where strong and weak models fail to contribute effectively. We define the collaboration as a strong-weak system. Through comprehensive experiments, we disclose a counterintuitive phenomenon the strong-weak system: a strong–weak collaboration may under-perform weak–weak combinations, revealing that cognitive mismatching are key bottlenecks limiting heterogeneous cooperation. To overcome these challenges, we propose an Entropy-Based Adaptive Guidance Framework that dynamically aligns the guidance with the cognitive state of each agent. The framework quantifies the understanding of weak agents’ through multi-dimensional entropy metrics—covering expression, uncertainty, structure, coherence, and relevance—and adaptively adjusts the intensity of the guidance at light, moderate and intensive levels. Furthermore, a Retrieval-Augmented Generation (RAG) mechanism is incorporated to retain successful collaboration experiences, enabling both immediate adaptation and long-term learning. Extensive experiments on three benchmark datasets, GSM8K , MBPP, and CVRP demonstrate that our approach consistently enhances the effectiveness and stability of heterogeneous collaboration. The results highlight that adaptive guidance not only mitigates cognitive imbalance but also establishes a scalable pathway toward more robust, cooperative multi-agent intelligence. 
% The source code can be found at: 
\end{abstract}

\section{Introduction}\textbf{}

\begin{figure}[t]
    \centering
    \includegraphics[width=\columnwidth]{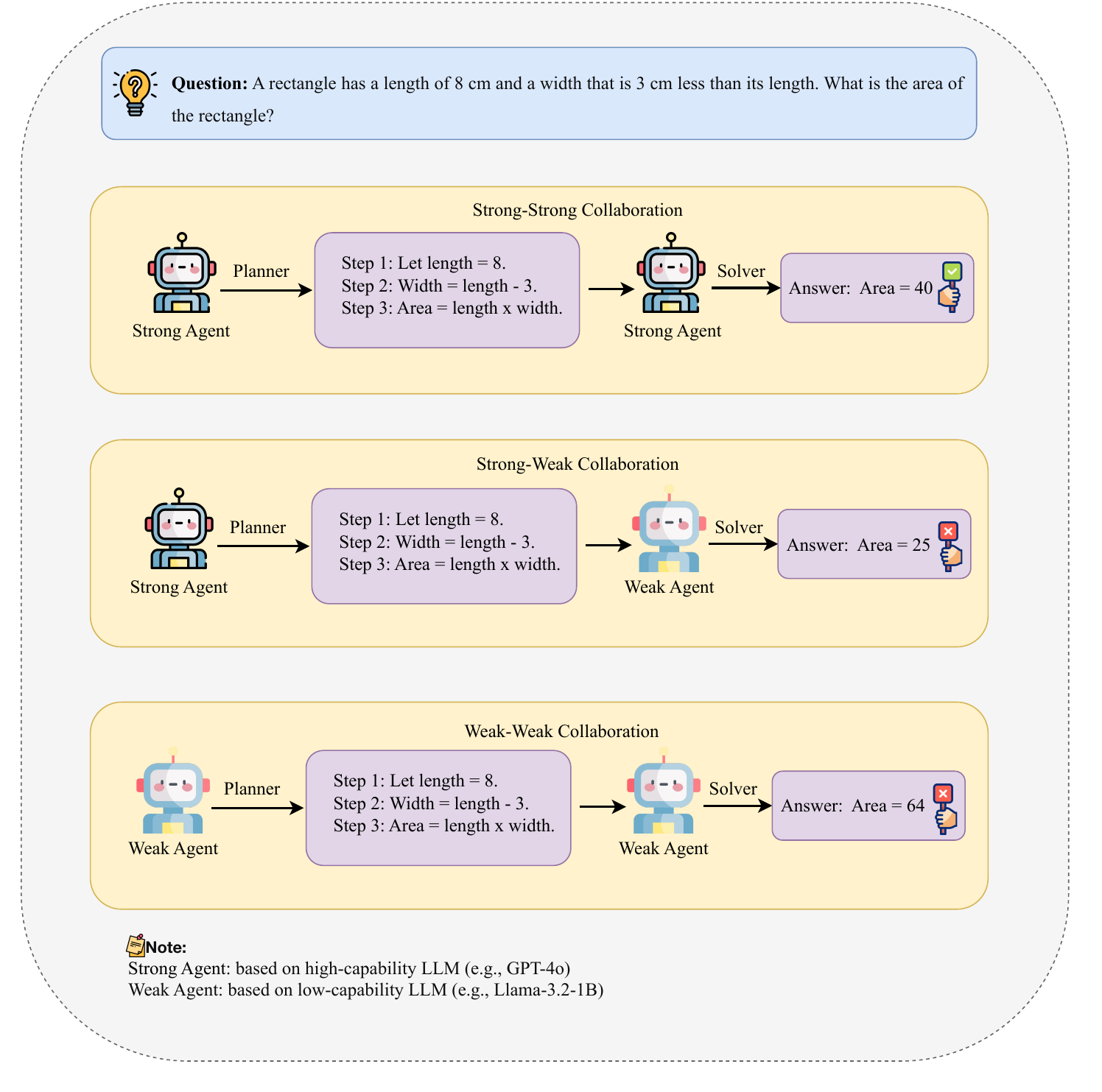}
    \caption{Illustration of collaboration patterns across three agent configurations on a simple mathematical reasoning task. \textbf{Top}: Strong-Strong collaboration achieves correct answer (Area = 40) through efficient information transfer. \textbf{Middle}: Strong-Weak collaboration produces incorrect answer (Area = 25) despite receiving identical high-quality guidance, demonstrating information loss due to cognitive mismatch. \textbf{Bottom}: Weak-Weak collaboration achieves better performance (Area = 64) than Strong-Weak through matched comprehension levels, revealing the negative synergy effect in heterogeneous collaboration.}
    \label{fig:collaboration_example}
\end{figure}

As large language models demonstrate remarkable capabilities in natural language understanding, knowledge reasoning, and complex task generation, artificial intelligence systems are evolving from single LLM-based agents to multi-agent systems with collaborative capabilities~\cite{li2024survey,he2025emerged,wang2025machine}. In this evolution, heterogeneous multi-agent systems (HMAS)~\cite{zhong2024heterogeneous}, collaborative systems composed of agents driven by LLMs with different capability levels, have attracted widespread attention due to their ability to integrate complementary strengths across models~\cite{guo2024large,ye2025x,yang2025autohma,wang2025unique}. In theory, strong agents can provide high-quality guidance and verification, while weak agents can execute specific tasks and learn from them. Their combination should produce significant synergistic effects, achieving performance beyond that of any single agent. However, empirical results often deviate significantly from this ideal assumption.

Existing research often overlooks a critical issue: in LLM-based heterogeneous agent systems, substantial differences in underlying model capabilities may lead to weak collaboration outcomes. Different LLMs vary greatly in reasoning ability, knowledge coverage, and expression accuracy. This imbalance in capability introduces new challenges to heterogeneous collaboration. To fully understand this issue, we constructed three typical collaboration configurations including Strong-Strong (SS), Weak-Weak (WW) and Strong-Weak (SW) setting to compare them across multiple tasks. The experimental results reveal a counterintuitive phenomenon: in some tasks, SW combinations perform worse than WW combinations. This indicates that high-quality suggestions from strong agents fail to transfer it ability to weak agents, resulting in excessive redundant interactions and information loss.

Further analysis shows that the root cause of this phenomenon lies in the asymmetry of information transfer and the mismatch in the comprehension cost. When weak agents receive complex reasoning chains generated by strong agents, they often face excessive cognitive overload, leading them to ignore, simplify, or misinterpret key information. These actions cause a decline in the effective rate of information transfer. This reveals that the performance of heterogeneous multi-agent collaboration depends not only on the upper-bound capability of the strongest agent in the system, but is more constrained by the bottleneck of weak agents in the system.

Although recent years have seen significant progress in multi-agent collaboration frameworks~\cite{wu2024autogen,hong2023metagpt,gao2402agentscope} and collaborative reasoning mechanisms~\cite{wei2022chain,liang2023encouraging,du2023improving}, these methods mainly target homogeneous or similarly capable agents and lack proactive guidance mechanisms to address capability differences in heterogeneous systems~\cite{sanwal2025layered,sani2024towards}. Existing methods such as Chain-of-Thought~\cite{wei2022chain}, multi-agent debate~\cite{liang2023encouraging}, and reflection mechanisms~\cite{chen2024magicore} perform well in homogeneous scenarios but show limited effectiveness in heterogeneous systems because they fail to adapt to the cognitive characteristics of weak agents. Therefore, how to break through performance bottlenecks in heterogeneous collaboration and enhance the understanding and execution capabilities of weak agents has become a core challenge that urgently needs to be addressed.

To this end, we propose an Entropy-Based Adaptive Guidance Framework that aims to enable strong agents to provide dynamic guidance matched to the cognitive characteristics of weak agents by quantifying their understanding states, thereby achieving efficient knowledge transfer and skill transmission. The framework contains three core components:
\begin{itemize}
\item Understanding Assessment via Entropy: We propose a multi-dimensional entropy measurement method to quantify the understanding level of weak agents. The system can accurately assess the cognitive state of weak agents on specific tasks.

\item Adaptive Guidance Strategies: Based on entropy assessment results, we design three guidance levels—light, moderate, and intensive, and introduce a dynamic threshold adjustment mechanism. This allows guidance intensity to adapt as weak agents learn, avoiding both over-intervention and cognitive overload.

\item Experience Retention with RAG: We introduce a Retrieval-Augmented Generation (RAG) module to record and reuse successful collaboration experiences. This combines immediate guidance with long-term knowledge accumulation.
\end{itemize}

The main contributions of our paper are as follows.
\begin{itemize}
\item We are the first to systematically identify and analyze the negative synergy effect in heterogeneous multi-agent systems, showing that weak-agent cognitive limitations form the main performance bottleneck.

\item We propose an entropy-based adaptive guidance framework that models the weak agent’s understanding dynamically and provides tiered guidance accordingly.

\item We integrate Retrieval-Augmented Generation (RAG) for structured experience retention, combining memory-based learning and real-time guidance to accelerate weak-agent improvement.

\item Extensive experiments on multiple benchmark tasks and agent configurations verify the effectiveness and robustness of our approach, offering a generalizable solution for heterogeneous agent collaboration.

\end{itemize}

\section{Related Works}
\subsection{Multi-Agent Systems with Heterogeneous LLMs}
With large language models (LLMs) demonstrating excellent abilities in natural language understanding, knowledge reasoning, and complex task generation, AI systems are gradually evolving from single LLM-based agents to multi-agent systems with collaborative capabilities. LLM-based multi-agent systems can be further divided into homogeneous and heterogeneous systems, depending on whether the agents rely on the same or different large language models~\cite{guo2024large}. In homogeneous systems, all agents use LLMs with the same structure and capability; in heterogeneous systems, agents are driven by different types of LLMs with diverse abilities, showing complementary strengths among models.

In question-answer scenarios, Ye et al.~\cite{ye2025x} proposed the heterogeneous LLM-driven multi-agent system paradigm X-MAS (Cross-Model Multi-Agent System). By integrating LLMs with different expertise, it significantly improves response quality and accuracy in multi-turn question answering. This system uses agent collaboration and division of labor to overcome the dependence on the capability of a single model in homogeneous systems, fully exploiting the unique strengths of each LLM in specific domains to achieve complementary abilities and better overall performance.
In physical environments, Yang et al.~\cite{yang2025autohma} proposed the AutoHMA-LLM framework with a hierarchical architecture that uses a cloud-based LLM as a central planner combined with device-specific LLMs and generative agents. This framework improves the efficiency and accuracy of task scheduling in complex and dynamic environments, demonstrating the collaborative potential of heterogeneous agents in real-world deployment.

\subsection{Collaboration Architectures in Multi-Agent Systems}
Multi-agent collaboration can be classified into three main types: centralized, distributed, and hybrid architectures~\cite{tran2025multi}. Each has unique designs and suits different application scenarios. Centralized collaboration relies on a central agent that gathers information, makes decisions, and assigns tasks. For example, AutoGen~\cite{wu2024autogen} centrally schedules multiple customizable LLM agents to efficiently manage dialogue and task allocation across domains. MetaGPT~\cite{hong2023metagpt} encodes standardized workflows as prompts and uses a pipeline role division to break down complex tasks, reducing errors and improving collaboration consistency. AgentScope~\cite{gao2402agentscope} uses message passing as its core, integrating tools and built-in agents with fault tolerance, multimodal processing, and flexible switching between local and distributed modes. MegaAgent~\cite{wang2024megaagent} builds a large autonomous system that dynamically generates agents based on task complexity without predefined workflows.

In task decomposition and collaboration, Ning et al.~\cite{ning2023skeleton} proposed Skeleton-of-Thought, which breaks down tasks into key points, enabling parallel collaboration that reduces delay and improves answer quality. AUTOACT~\cite{qiao2024autoact} uses centralized planning to coordinate sub-agents in question answering. Suzgun et al.~\cite{suzgun2024meta} developed meta-prompting to turn a single language model into a manager that integrates multiple independent queries, allowing effective expert collaboration.

Distributed collaboration focuses on agents making decisions locally and coordinating globally through direct interaction. Liang et al.~\cite{liang2023encouraging} introduced MAD, where multiple independent agents debate their views in parallel. AutoAgents~\cite{chen2023autoagents} create specialized agents dynamically to complete tasks efficiently without a central controller. Jeyakumar et al.~\cite{jeyakumar2024advancing} proposed a framework using dynamic task graphs and automatic tool selection for asynchronous and efficient collaboration.

Regarding system design, AgentNet~\cite{yang2025agentnet} combines retrieval-augmented generation with evolving network structures and autonomous specialization to build an efficient, fully decentralized system. DecentLLMs~\cite{jo2025byzantin} uses multiple working agents to generate answers and multiple evaluating agents to score them independently. This decentralized consensus approach improves security and decision quality beyond traditional leader-based coordination.

Hybrid collaboration mixes global guidance with local autonomy to balance coordination and flexibility. Li et al.~\cite{li2023camel} proposed Role-Playing, guiding multi-agent interactions with role-play and heuristic prompts, aligning agents with human intentions while allowing autonomy. ChatEval~\cite{chan2023chateval} uses multiple LLM agents with distinct roles to debate and jointly evaluate generated text, demonstrating benefits of multi-role collaboration. DyLAN~\cite{liu2024dynamic} applies a two-step approach: centralized optimization based on agent importance scores, followed by autonomous collaboration within a dynamic communication network, blending central control and distributed autonomy effectively.

\subsection{Collaborative Reasoning Mechanisms in Multi-Agent Systems}
As the complexity of multi-agent systems and task difficulty increase, relying on independent reasoning by a single agent is no longer sufficient. Collaborative reasoning mechanisms have thus become crucial to improve overall performance. Multi-agent collaborative reasoning aims to achieve collective intelligence stronger than any single agent by sharing information, critical reflection, and iterative optimization. Chain-of-Thought (CoT) reasoning guides agents to generate a series of intermediate reasoning steps, effectively enhancing performance on complex tasks. Wei et al.~\cite{wei2022chain} demonstrated that a small number of CoT examples can trigger powerful reasoning abilities in large language models. Building on this, Sanwal et al.~\cite{sanwal2025layered} proposed Layered-CoT, which divides the reasoning process into multiple levels, each supported by external verification and user feedback to improve reliability. Sani et al.~\cite{sani2024towards} employed multi-agent collaboration to break down complex reasoning into hierarchically verifiable steps, significantly increasing transparency and accuracy. Low et al.~\cite{low2025surgraw} introduced the SurgRAW framework in robot-assisted surgery, combining multi-agent collaboration with layered CoT to achieve structured and domain-aware reasoning, greatly reducing hallucinations and enhancing reasoning clarity and precision.

Debate and argumentation mechanisms promote deeper analysis by having agents present and challenge each other’s viewpoints. For example, Du et al.\cite{du2023improving} proposed a system where multiple language model instances engage in multi-round debates and reasoning exchanges, effectively reducing errors and hallucinations while improving the factuality and reliability of generated content. Liang et al.\cite{liang2023encouraging} introduced the Multi-Agent Debate (MAD) framework, where several agents engage in “tit-for-tat” debates and a judge integrates the arguments into a final decision. This approach effectively mitigates the Degeneration-of-Thought problem observed in single-agent self-reflection, significantly improving performance on complex reasoning tasks, while emphasizing the importance of impartial judging and model diversity. Zhang et al.~\cite{zhang2025if} further proposed Heter-MAD, enabling a single large language model to access outputs from multiple heterogeneous foundation models, which further improves the performance and robustness of multi-agent debate systems.

Reflection and iterative refinement mechanisms encourage agents to self-check and revise their previous reasoning results. Through multiple iterations, these mechanisms improve reasoning quality, correct early mistakes, and gradually converge to better solutions. Chen et al.\cite{chen2024magicore} proposed the MAGICORE framework, which dynamically chooses between coarse-grained aggregation and fine-grained multi-agent iterative refinement based on problem difficulty. It uses external reward models to precisely localize errors and employs a cyclic interaction among Solver, Reviewer, and Refiner agents to significantly enhance reasoning performance. Yuan et al.\cite{yuan2025agent} introduced Agent-R, an iterative self-training framework based on Monte Carlo Tree Search (MCTS) that focuses on detecting and correcting errors in real time during agent execution, improving error recovery and learning efficiency. Low et al.~\cite{low2025mirror} presented the MIRROR framework, combining intra-reflection (before action) and inter-reflection (after action), simulating human thinking processes before and after decisions. This design effectively prevents error propagation and promptly corrects mistakes during execution, substantially improving robustness and performance in complex multi-agent tasks.

\subsection{Research Gap}

Although heterogeneous multi-agent systems are increasingly applied to complex tasks, significant gaps remain in current research. First, the issue of uneven capabilities among individual agents in heterogeneous large model-based multi-agent systems is often overlooked. Due to differences in ability and performance, agents driven by weaker large models frequently become bottlenecks in system collaboration, reducing task efficiency and result quality. For example, in intelligent manufacturing factories~\cite{bahrpeyma2022review}, different agents handle various stages of production; if some weaker agents perform poorly in material inspection or fault diagnosis, the overall efficiency of the production line and the quality of the final product will be directly affected. 

However, existing literature rarely provides a systematic analysis of this problem~\cite{song2025digital, ye2025x}. Second, there is a lack of strategies to improve weaker agents in heterogeneous multi-agent systems. Current methods mainly focus on cooperation between agents with similar abilities and lack active guidance mechanisms tailored to ability differences, limiting the full potential of the agent group~\cite{liang2023encouraging, sanwal2025layered}. Finally, due to differences in capabilities, knowledge, and resources, heterogeneous agents face greater challenges in information sharing and collaborative reasoning. Most existing collaborative reasoning approaches are designed for homogeneous or similarly capable agents and are not directly applicable to heterogeneous environments~\cite{sani2024towards, low2025surgraw}. In summary, addressing collaboration bottlenecks caused by uneven model capabilities in heterogeneous multi-agent systems and enhancing weaker agents to improve overall collaboration efficiency remain key challenges in this field.

\section{Preliminaries}
\subsection{Agent Definition}
In multi-agent systems and artificial intelligence, a single agent typically refers to an autonomous entity that can perceive, make decisions, and take actions within a specific environment. Formally, an LLM-based agent can be defined as $ \mathcal{A} = (M,  S, A, \pi)$
% \begin{equation}
%      \mathcal{A} = (M,  S, A, \pi)
% \end{equation}
The meaning of each component is as follows:
\begin{enumerate}
\item[•] $M$ represents the structure of the core model that supports the operation of the agent, that is, the large language model that drives the agent’s behavior. It forms the basis for agent reasoning and decision-making.
\item[•] $S$ represents the internal state of the agent, reflecting its current knowledge and perception of the environment. This includes the contextual state, the knowledge state, and the cognitive state.
\item[•] $A$ represents the set of executable actions available to the agent and defines how it interacts with the environment and other agents.
\item[•] $\pi$ represents the agent’s decision policy, which determines how to select the appropriate action given the current state.
\end{enumerate}
An LLM-based agent repeatedly executes a perception–decision–action cycle, continuously interacting with the environment and other agents to achieve its goals.

\subsection{Heterogeneous Multi-Agent Systems}
A multi-agent system consists of multiple intelligent agents that can interact, collaborate, or compete with each other to complete complex tasks that are difficult for a single agent to accomplish independently~\cite{}. A multi-agent system can be formally defined as $ MAS=(\mathcal{A},  \mathcal{E},   \mathcal{C},  \mathcal{T})$
% \begin{equation}
%      MAS=(\mathcal{A},  \mathcal{E},   \mathcal{C},  \mathcal{T})
% \end{equation}

\begin{enumerate}
\item[•] $\mathcal{A}$ represents a collection of LLM-based agents, where $i$ denotes the number of agents. The agents can be dynamically adjusted according to the current needs of the system.
\item[•] $\mathcal{E}$ represents the external environment where agents interact and operate, which can be either a physical environment or a virtual environment.
\item[•] $ \mathcal{C}$ defines the methods and rules for information exchange between agents, such as communication channels or message formats.
\item[•] $\mathcal{T}$ represents the set of tasks that the system needs to complete through collaboration. Tasks may be completed independently by a single agent or may require collaboration among multiple agents.
\end{enumerate}
In a Heterogeneous Multi-Agent System (H-MAS), the agents in set $\mathcal{A}$ are not uniform, but are built based on different large language model instances with different capabilities, knowledge scope, and reasoning levels~\cite{ ye2025x}. This heterogeneity may come from differences in model architecture, parameter size, training data distribution, or domain expertise.
\begin{definition}[Heterogeneous Multi-Agent System]
A heterogeneous multi-agent system is a framework comprising multiple agents that are built upon different large language models and collaborate to accomplish complex tasks. A heterogeneous multi-agent system can be represented as $HMAS=(\mathcal{A}_{H}, \mathcal{E},  \mathcal{C},  \mathcal{T})$
\end{definition}
Where $\mathcal{A}_{H}=\{a^{M_1}_{1}, a^{M_2}_{2},...,a^{M_i}_{i}\}$ represents a heterogeneous set of agents, $M=\{M_1, M_2,...,M_i\}$ denotes the set of large language models, $a^{M_i}_{i}$ represents an agent based on large language model $M_i$.

Heterogeneity constraint, $\exists j,k \in \{1,2,\ldots,i\}, \, j\neq k: \, M_j \neq M_k$, the system must have at least two agents that use different underlying large language models.

In a heterogeneous multi-agent system, the capability of an agent mainly depends on the performance of its underlying large language model. Therefore, we classify agents according to the strength of their underlying model.

Strong agents $\mathcal{A}_{strong}$  are built on high-performance large language models, with strong reasoning ability and broad knowledge coverage. They can independently make high-quality decisions or provide solutions for complex tasks.

Weak agents $\mathcal{A}_{weak}$ are built on large language models with relatively limited capabilities, having restricted reasoning ability and narrower knowledge coverage. They can independently handle simple and straightforward tasks.

\subsection{Information Entropy for Understanding Assessment}
Information entropy~\cite{gray2011entropy} is used to measure the uncertainty or amount of information of a random variable. For a discrete random variable $X$, its information entropy is defined as.
\begin{equation}
H(X) = -\sum_{i=1}^{n} p_i log_2 p_i
\end{equation}
Where $p_i$ is the probability that the random variable $X$ takes its $i\-$th value, that is, $p_i = P(X = x_i)$.

The main properties of information entropy are as follows:
\begin{enumerate}
    \item[•] Non-negativity: For any random variable $X$, $H(X) \geq 0$, and the entropy is zero when the distribution is completely certain.
    \item[•] Upper bound: $H(X) \leq \log_2 n$, where the maximum value is reached when $X$ follows a uniform distribution.
    \item[•] Additivity: For independent random variables $X$ and $Y$, $H(X, Y) = H(X) + H(Y)$.
\end{enumerate}

In LLM-based agents, generating text or actions is usually a process based on a probability distribution $P(y|x)$, where $x$ is the current context or state, and $y$ is a candidate output.
The entropy of this probability distribution can be defined as:
\begin{equation}
H(Y|X=x) = -\sum_{y \in Y} P(y|x) log_2 P(y|x)
\end{equation}
Where $Y$ is the set of all possible outputs. A higher entropy means that the model is more uncertain in choosing its output, while a lower entropy means the model is highly confident about certain outputs.

In an agent reasoning and decision-making, output uncertainty can be seen as a measure of the agent’s understanding of a problem. We assume that the agent’s understanding level $U$ has a monotonic decreasing relationship with the entropy $H$ of its output probability distribution, which can be expressed as:
\begin{equation}
U = f(H),  \frac{\text{d}U}{\text{d}H}  < 0
\end{equation}
where a higher entropy indicates a lower level of understanding, and a lower entropy indicates a higher level of understanding.

\begin{figure*}[ht!]
    \centering
    \includegraphics[width=0.95\textwidth]{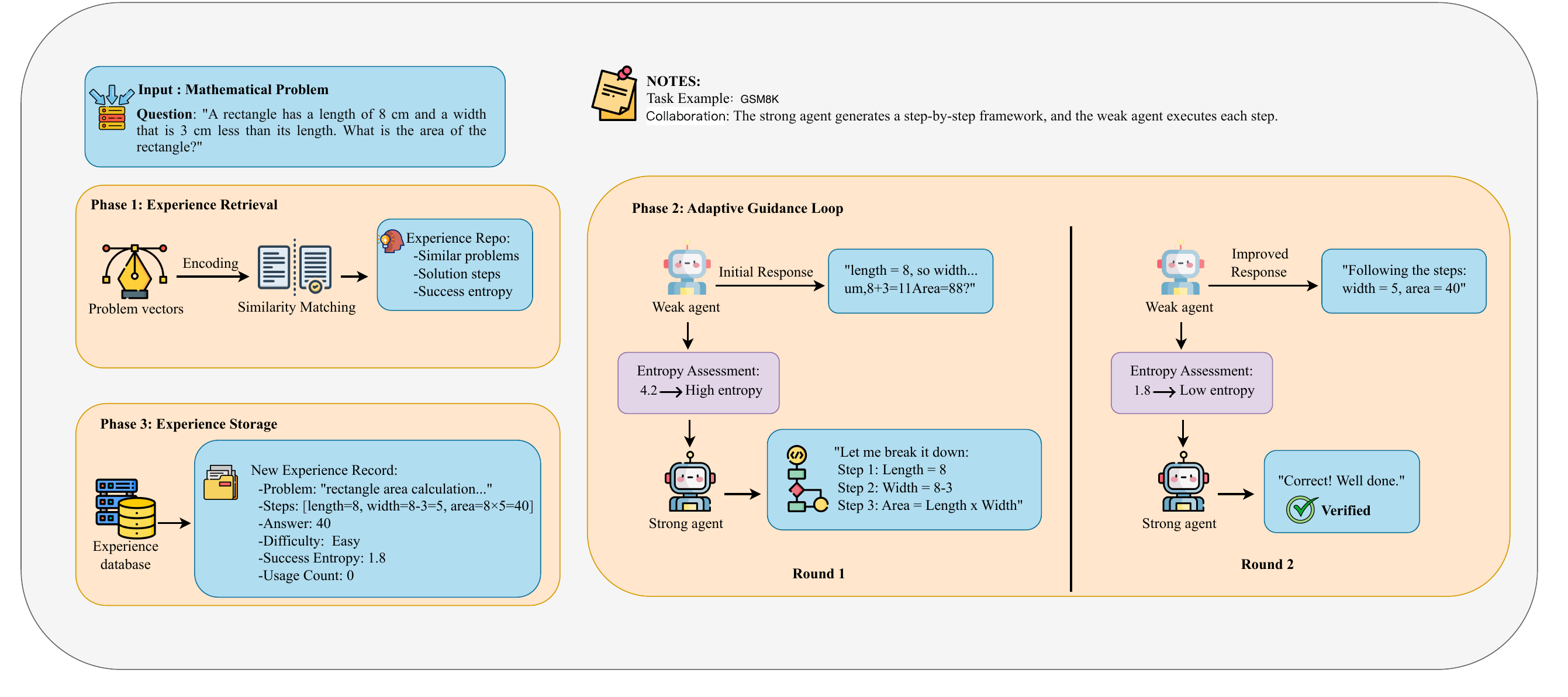}
    \caption{
        Overall framework of the heterogeneous multi-agent collaboration process.
    }
    \label{fig:framework}
\end{figure*}
\section{Understanding Collaboration Bottlenecks in Heterogeneous Multi-Agent Systems}
\subsection{Research Motivation}\label{sec:4.1}
In heterogeneous multi-agent systems, agents with different capabilities can collaborate to achieve complementary advantages, leading to performance beyond that of any single agent. In theory, strong agents can provide high-quality guidance and verification, while weak agents can carry out specific tasks and learn from them. Their combination should, in principle, produce a significant synergistic effect.

However, existing studies often overlook a key factor: in heterogeneous agent systems based on large language models, the substantial imbalance in underlying model capabilities may result in weaker-than-expected collaboration. Different large language models vary greatly in reasoning ability, knowledge coverage, and expression accuracy, which introduces unprecedented challenges to heterogeneous cooperation.

To better understand this issue, we conducted preliminary experiments by constructing three types of collaboration settings for the same task: strong agent–strong agent collaboration, weak agent–weak agent collaboration, and heterogeneous collaboration between a strong and a weak agent. The results revealed a puzzling phenomenon: in some cases, the performance of strong–weak collaboration was even worse than that of two weak agents working together. This suggests that the high-quality suggestions from the strong agent are not effectively transferred to the weak agent. As a result, many inefficient interactions occur during collaboration, and the actual outcome of heterogeneous cooperation falls far short of theoretical expectations.

To illustrate this phenomenon in concrete terms, consider the example shown in Figure~\ref{fig:collaboration_example}. Given a simple mathematical reasoning problem, "A rectangle has a length of 8 cm and a width that is 3 cm less than its length. What is the area?"—three collaboration configurations produce strikingly different outcomes. In the Strong-Strong collaboration (top), both agents successfully follow the structured reasoning steps to arrive at the correct answer (Area = 40 cm²). The planning agent provides a clear three-step framework, and the solver agent executes it accurately. 

In contrast, the Strong-Weak collaboration (middle) receives identical high-quality guidance from the strong planning agent, but produces an incorrect answer (Area = 25 cm²). This failure occurs despite the weak agent having access to the same clear reasoning steps. Analysis reveals that the weak agent misinterprets "Width = length - 3" during execution, demonstrating the cognitive overload problem: the guidance is understood superficially but not executed correctly. 

Most surprisingly, the Weak-Weak collaboration (bottom) achieves a different result (Area = 64 cm²). While also incorrect, this configuration shows that when both agents operate at similar cognitive levels, they can at least maintain consistent reasoning patterns without the confusion introduced by mismatched comprehension capabilities. This simple example vividly demonstrates the negative synergy effect, which adding a strong agent's guidance to a weak agent can actually worsen performance compared to collaboration between agents of similar capabilities.

Based on these observations, we propose the following research questions:
\begin{enumerate}
\item[•] In heterogeneous multi-agent systems, what factors become the key bottlenecks for collaboration effectiveness?
\item[•] Why are high-quality suggestions from strong agents not effectively transferred to weak agents?
\item[•] What performance patterns emerge when agents with different capability levels collaborate?
\end{enumerate}
Furthermore, we present the following core research hypotheses:
\begin{enumerate}
\item[•] In heterogeneous multi-agent collaboration, the capability level of the weak agents is the critical bottleneck that determines overall performance.
\item[•] Weak agents may experience cognitive overload when processing the complex guidance provided by strong agents. This overload can reduce information processing efficiency and lead to weaker collaboration outcomes compared to agents with similar capability levels.
\item[•] The traditional guidance strategies of strong agents do not align well with the cognitive characteristics of weak agents. More adaptive guidance mechanisms are needed to achieve effective knowledge and skill transfer.
\end{enumerate}

\subsection{Exploring Collaboration Bottlenecks in Heterogeneous MAS}
To systematically investigate these potential bottlenecks, we designed an exploratory experiment to quantify the actual performance of agent collaboration under different capability configurations. We carefully selected three complementary benchmark tasks to comprehensively evaluate collaborative performance: GSM8K (mathematical reasoning), MBPP (code generation), and CVRP (vehicle routing planning). These tasks represent distinct cognitive challenges including symbolic reasoning, program synthesis, and combinatorial optimization, ensuring the generalizability of our experimental results.

For agent configuration, we constructed a heterogeneous agent pool comprising four mainstream large language models. Based on significant differences in model capabilities, we categorized GPT-4o and Claude-3.5 as "strong agents," while smaller-scale Qwen-2.5-1B and LLaMA-3.2-3B were defined as "weak agents." Building on this classification, we designed three typical collaboration configurations: Strong-Strong (SS), Weak-Weak (WW), and our primary focus, Strong-Weak (SW) combinations, thereby establishing a comprehensive comparative framework.

\begin{figure*}[ht!]
    \centering
    \includegraphics[width=0.95\textwidth]{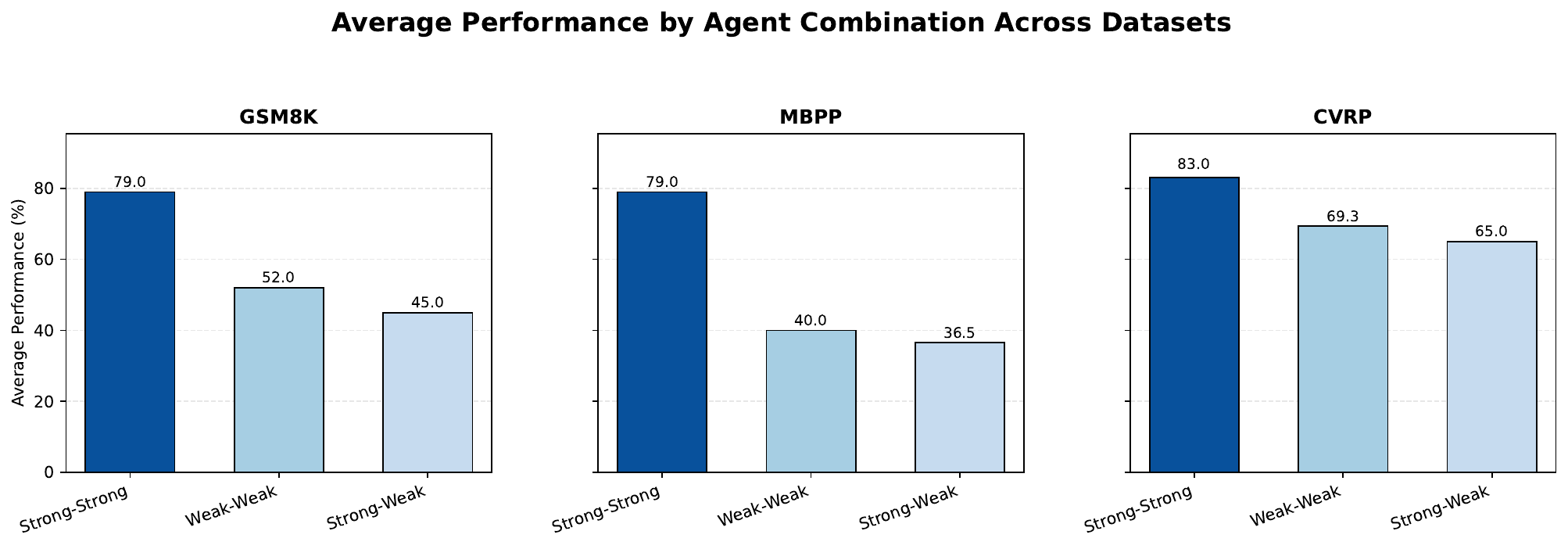}
    \caption{Average performance of heterogeneous agent combinations (Strong-Strong, Weak-Weak, Strong-Weak) across three benchmark datasets (GSM8K, MBPP, CVRP). Each subplot corresponds to a dataset.}
    \label{fig:agent_combinations}
\end{figure*}

As shown in Figure~\ref{fig:agent_combinations}, the experimental results reveal significant performance disparities in heterogeneous collaboration. Across all three benchmark tasks, the SS combination consistently achieved optimal performance with average accuracies of 79.0\%, 79.0\%, and 83.0\% on GSM8K, MBPP, and CVRP respectively, validating the natural advantages of collaboration between high-capability agents.

The combination of WW demonstrated moderate performance levels (GSM8K: 52.0\%, MBPP: 40.0\%, CVRP: 69.3\%), providing an important baseline for evaluating the effectiveness of heterogeneous collaboration. In particular, the SW combination performed unexpectedly below expectations—its performance in GSM8K (45.0\%) and MBPP (36.5\%) even fell below the WW combination, only approaching the WW level in the CVRP task (65.0\%). This performance inversion is particularly striking: compared to the WW combination, the SW combination showed a 7 percentage point decrease on GSM8K and a 3.5 percentage point decrease on MBPP. This counterintuitive result strongly suggests that in heterogeneous collaboration scenarios, the knowledge and reasoning capabilities of strong agents not only fail to enhance overall performance, but may actually reduce efficiency due to collaboration overhead and cognitive mismatch. This "negative synergy effect" directly points to the core bottleneck in heterogeneous multi-agent systems—the cognitive limitations of weak agents severely hinder the absorption and utilization of high-quality knowledge.

These findings not only confirm the existence of significant performance bottlenecks in heterogeneous collaboration, but also provide clear direction for our subsequent design of targeted guidance strategies: how to make strong agents'guidance better adapt to weak agents' cognitive characteristics will be key to breaking through heterogeneous collaboration bottlenecks.

% To test the research hypotheses proposed in Section~\ref{sec:4.1}, especially the weak-agent bottleneck hypothesis, we designed systematic comparative experiments to analyze the collaboration performance under different agent configurations. The goal of these experiments is to identify the key bottleneck factors in heterogeneous multi-agent collaboration.
% \subsubsection{Experimental Setup}
% Dataset and model. In this study, we randomly sampled 50 questions from the GSM8K mathematical reasoning dataset. The sampling was stratified by difficulty level with the following ratio: Easy (30\%), Medium (50\%), and Hard (20\%). We selected four models: GPT-4o, Claude-3.5, Qwen-2.5-1B, and LLaMA-3.2-3B. Among them, GPT-4o and Claude-3.5 were considered strong agents, while Qwen-2.5-1B and LLaMA-3.2-3B were considered weak agents.

% Collaboration Configurations. We designed three collaboration configurations (strong–strong, weak–weak, and strong–weak) to form a comparison system. For each configuration, we tested two collaboration modes:

% Solve–Review Mode: one agent solves the problem, while the other reviews and corrects the answer.

% Framework–Execution Mode: one agent provides a solution framework, while the other follows the framework to complete the reasoning step by step.

% Evaluation Metric. The main evaluation metric for model performance was accuracy.

% \subsubsection{Overall performance comparison}

\subsection{Analysis and Implications of Collaboration Bottlenecks}
The above experimental results clearly validate our core hypothesis: in heterogeneous multi-agent collaboration, weak agents indeed form the critical performance bottleneck. The average accuracy of strong–weak collaboration is comparable to that of weak–weak collaboration, and in some cases, it is even lower.

The main reason for this performance decline lies in the asymmetry of information transfer and the mismatch in comprehension cost. When receiving and interpreting the highly complex reasoning chains produced by strong agents, weak agents often face excessive cognitive load. This leads them to ignore, simplify, or misinterpret key information during task execution. In addition, strong agents may lack adaptive strategies that account for the capability limitations of weak agents. As a result, the number of interaction rounds increases, but the effective information transfer rate decreases.

This finding shows that in heterogeneous multi-agent systems, collaboration effectiveness is not only determined by the upper-bound capabilities of the agents but also constrained by the weak agents’ ability to absorb and execute information. Therefore, the key to improving heterogeneous collaboration lies in designing more adaptive guidance mechanisms. Such mechanisms should enable strong agents to convey high-quality information in forms that weak agents can understand and act upon, thereby overcoming the collaboration bottleneck.

\section{Method: Entropy-Based Adaptive Guidance for Heterogeneous Agents}
\subsection{Overall Framework}
We propose a framework for the collaboration of adaptive guidance based on information entropy. This approach quantifies the understanding state of weak agents, enabling strong agents to provide adaptive guidance. It also incorporates a retrieval-augmented generation mechanism to leverage historical experience.

As shown in Figure~\ref{fig:framework}, the framework consists of three phases: experience retrieval, adaptive guidance loop, and RAG-based experience storage. Figure~\ref{fig:framework} illustrates the complete collaboration process. Using a mathematical reasoning task as an example, the framework operates through three phases: 

\textbf{Phase 1: Retrieval of} experiences. The system encodes the input problem and retrieves relevant historical successful cases from the experience repository, providing initial reference for the weak agent. 

\textbf{Phase 2: Adaptive Guidance Loop}. This is the core of the framework. In Round 1, the weak agent's initial response shows confusion ("um, 8+3=11?"). The system evaluates this and obtains high understanding entropy ($H_u = 4.2 > \tau_2$), which triggers intensive guidance mode. The problem is decomposed into detailed sub-steps. In Round 2, the weak agent's improved response shows that understanding entropy has decreased to 1.8. The system then switches to light guidance, providing only brief verification. This dynamic adjustment reflects the framework's adaptability—precisely matching guidance intensity to the understanding state.

\textbf{Phase 3: Experience Storage}. After successful collaboration, the system packages the problem, solution steps, answer, difficulty level, and success entropy into the experience repository, forming a continuous learning cycle. Unlike traditional Chain-of-Thought methods, our framework quantifies understanding states through multi-dimensional entropy metrics, dynamically selects guidance levels based on understanding entropy, and progressively reduces intervention through dynamic threshold adjustment.

\subsection{Understanding Assessment via Entropy}
To enable adaptive guidance in heterogeneous multi-agent systems, we propose an entropy-based understanding assessment mechanism. This mechanism quantifies the uncertainty and reliability of agents during collaborative reasoning. By computing the response entropy of an agent on a specific task, it measures the agent’s level of understanding and provides the basis for subsequent hierarchical guidance.

We define understanding entropy $H_u(R|P, \mathcal{T}) $ as a quantitative indicator of an agent’s uncertainty in understanding a problem $P$ under a given task type $T$:
\begin{equation}
H_u(R|P, \mathcal{T}) = \mathcal{F}_{\mathcal{T}}(R, P)
\end{equation}
Here, $F_T$ denotes the feature extraction function associated with task type $T$, $H_u(R|P, \mathcal{T}) \in [0, +\infty]$. The degree of understanding is inversely related to entropy, higher entropy indicates lower understanding, while lower entropy indicates higher understanding.

Given a problem $P$ and the agent’s response $R$, we define the response entropy as a weighted combination of multiple factors:
\begin{equation}
\begin{split}
H_u(R|P, \mathcal{T}) = & \ H_{\text{expression}}(R) 
   + H_{\text{uncertainty}}(R, T) + H_{\text{structure}}(R) \\
   & - H_{\text{coherence}}(R, T) 
   - H_{\text{relevance}}(R, P, T)
\end{split}
\end{equation}
where higher entropy corresponds to lower understanding, whereas lower entropy reflects higher understanding.

Specifically, the meanings of the five types of entropy are as follows:

$H_{\text{expression}}(R) = - \sum_{W_i \in V(R)} p(w_iR|)log_2 p(w_i|R)$ measures the dispersion of word usage. The more evenly distributed the vocabulary, the higher the entropy, which indicates greater uncertainty at the language level.

$H_{\text{uncertainty}}(R, T)= \sum_{{pattern} \in U_T} w_{pattern}$ captures semantic uncertainty beyond probability-level variation. We construct an uncertainty lexicon (e.g., possible, maybe, uncertain) to identify task-related uncertainty expressions. The higher the weight of uncertainty words in the response, the higher the uncertainty entropy.

$H_{\text{structure}}(R)$ evaluates the information density with respect to response length. If the response includes logical and procedural markers (e.g., first, then, because, therefore), it reflects a reasoning structure and reduces entropy.

$H_{\text{coherence}}(R, T)$ identifies task-specific logical connection patterns, ensuring that reasoning steps are consistent and well-structured.

$H_{\text{relevance}}(R, P, T)$ evaluates the alignment of the response with the problem in task-specific dimensions. Both overly short and overly long responses are penalized, so that the entropy evaluation reflects the appropriateness of the answer.

To reflect the varying importance of these entropy indicators across different task types, we design a task-dependent weight matrix:
\begin{equation}
\begin{split}
\mathbf{W}^T = \begin{bmatrix} w_\text{expression}^T \\ w_\text{uncertainty}^T \\ w_\text{structure}^T \\ w_\text{coherence}^T \\ w_\text{relevance}^T \end{bmatrix}
\end{split}
\end{equation}
By multiplying each entropy component with this weight matrix, we obtain a weighted overall response entropy, which provides a more accurate measurement of the agent’s understanding in a given task.

\subsection{Adaptive Guidance Strategies}
Based on the understanding assessment mechanism introduced in the previous section, we design an adaptive guidance strategy. The goal is to dynamically adjust the guidance intensity of strong agents according to the understanding level of weak agents, thereby improving the efficiency of heterogeneous multi-agent collaboration.

To better match the cognitive state of weak agents, the guidance strategy is divided into three levels:
\begin{equation}
G_{\text{level}}(H_u, T, t) =
\begin{cases}
G_{\text{light}}(P, R, T), & \text{if } H_u \leq \tau_1^{T}(t), \\[6pt]
G_{\text{moderate}}(P, R, T), & \text{if } \tau_1^{T}(t) < H_u \leq \tau_2^{T}(t), \\[6pt]
G_{\text{intensive}}(P, R, T), & \text{if } H_u > \tau_2^{T}(t).
\end{cases}
\end{equation}
Here, $G_{\text{light}}$, $G_{\text{moderate}}$, and $G_{\text{intensive}}$ represent the light, moderate, and intensive guidance strategies, respectively. The parameters $\tau_1^T(t)$ and $\tau_2^T(t)$ denote the dynamic thresholds for task $T$ at time $t$.

When the response entropy of the weak agent satisfies $H_u(R|P,T) \leq \tau_1$, it indicates that the agent has sufficient understanding of the problem and possesses the basic ability to complete the task. In this case, the strong agent only provides minor verification or correction suggestions. This reduces unnecessary intervention, preserves the autonomy of the weak agent, and ensures its ability to complete the task independently.
\begin{equation}
G_{\text{light}}(P,R,T)=V_{\text{verification}}(R)+C_{\text{correction}}(R)+E_{\text{encouragement}} 
\end{equation}

When the response entropy lies in the range $\tau_1 < H_u(R|P,T) \leq \tau_2$, it shows that the weak agent has a partial understanding of the problem but still faces uncertainty. In this case, the strong agent provides conceptual guidance, while the weak agent carries out the execution. This allows necessary support while maintaining autonomy.
\begin{equation}
G_{\text{moderate}} (P,R,T)=D_{\text{diagnosis}} (R,P)+F_{\text{framework}}^{T} (P)+S_{\text{suggestion}} (R)
\end{equation}

When the response entropy exceeds $\tau_2$, it indicates that the weak agent lacks sufficient understanding and faces significant cognitive difficulty. In this case, the strong agent adopts an intensive step-by-step guidance strategy, decomposing the complex problem into manageable sub-tasks. By offering concrete examples and strategy suggestions, the strong agent gradually reduces the uncertainty of the weak agent until it can complete the task independently.
\begin{equation}
G_{\text{intensive}} (P,R,T)=E_{\text{analysis}} (R)+R_{\text{reconstruction}}^T (P)+G_{\text{guidance}} (P,R)
\end{equation}

In the multi-round interactions of heterogeneous multi-agent collaboration, weak agents gradually improve their task understanding by receiving guidance from strong agents. This learning process can be described as the dynamic evolution of the weak agent’s cognitive state:
\begin{equation}
U_{w}^{(t+1)} = U_{w}^{(t)} + \Delta U\big(G^{(t)}, H_{u}^{(t)}\big)
\end{equation}
Where, $U_{w}^{(t)}$ denotes the understanding level of the weak agent after round $t$, $G^{(t)}$ denotes the guidance provided by the strong agent in round $t$, and $\Delta U(\cdot)$ represents the improvement in understanding brought by the guidance.

To enhance the flexibility and adaptability of the guidance strategy, we propose a dynamic threshold adjustment mechanism based on multi-round interactions. Let $\tau_1$ and $\tau_2$ represent the thresholds for light and moderate guidance, respectively. In each interaction round r, the weak agent generates a response, and its understanding level is evaluated by the response entropy $H_r$. The thresholds are then updated according to the agent’s progress, so that the guidance intensity adaptively decreases as the weak agent’s capability improves. This prevents over-intervention and avoids cognitive overload.

To avoid excessive decreases of the thresholds, we introduce a saturation-protected adjustment function:
\begin{equation}
A(r) = \min \big( \lambda \cdot (r - 1), A_{\max} \big)
\end{equation}
where $\lambda = 0.2$ is the learning rate, and $A_{\max} = 0.6$ is the maximum adjustment value.

The dynamic threshold update rules are defined as:

\begin{equation}
\begin{aligned}
\tau_1(r) &= \max \big( \tau_{1,\text{base}} - A(r), \, \tau_{1,\min} \big), \\
\tau_2(r) &= \max \big( \tau_{2,\text{base}} - A(r), \, \tau_{2,\min} \big)
\end{aligned}
\end{equation}
where $\tau_{1,\text{base}} = 2.0$ and $\tau_{2,\text{base}} = 3.5$ are the base thresholds, and $\tau_{1,\min} = 1.0$ and $\tau_{2,\min} = 2.0$ are the lower bounds for protection.

\subsection{Experience Retention with RAG}
In heterogeneous multi-agent systems, the successful experiences accumulated by weak agents during multi-round interactions serve as key resources for improving their capabilities. To effectively preserve and reuse these experiences, we introduce an experience management mechanism based on retrieval-augmented generation (RAG). This mechanism consists of two core modules: experience storage and experience retrieval and application.

\textbf{Experience Storage}. After each successful collaboration, the system packages task-related information into a structured experience record $E$ and stores it in the experience repository. Each experience record is defined as:
\begin{equation}
E = \langle P_{\text{text}}, S_{\text{steps}}, A_{\text{final}}, D_{\text{level}}, H_{\text{success}}, U_{\text{count}} \rangle
\end{equation}
where the components are defined as follows:
\begin{enumerate}
    \item[•] $P_{\text{text}}$: the text description of the problem, used for similarity matching in later retrieval.
     \item[•] $S_{\text{steps}}$: the sequence of successful solution steps, containing the complete reasoning process.
     \item[•] $A_{\text{final}}$: the final correct answer, used to verify the validity of the solution.
     \item[•] $D_{\text{level}}$: the difficulty level of the problem (easy / medium / hard), used for difficulty-based matching.
     \item[•] $H_{\text{success}}$: the understanding entropy at the time of successful collaboration, reflecting the quality of problem solving.
      \item[•] $U_{\text{count}}$: the usage count of the experience, recording how often the experience has been reused.
\end{enumerate}

\textbf{Experience Retrieval and Application}. When a weak agent processes a new problem, the system first encodes the problem into a vector representation and then retrieves relevant experiences from the experience repository.

The current problem $P_{\text{current}}$ is encoded using the TF-IDF method:
\begin{equation}
\mathbf{v}_P = [w_1, w_2, \dots, w_d], \quad w_i = tf_i \cdot \log \frac{N}{df_i}
\end{equation}
where $tf_i$ denotes the frequency of term i in the problem, $df_i$ represents the number of experience records containing term $i$, and N is the total number of problems in the experience repository. This representation effectively captures the semantic features of the problem and provides a basis for similarity computation in retrieval.

Using this retrieval-augmented generation mechanism, weak agents can quickly leverage historical successful experiences when facing new tasks. This process reduces the initial understanding entropy, lowers the exploration cost, and enables the generation of higher-quality responses, thereby accelerating the improvement of the weak agent’s cognitive capabilities.

Algorithm~\ref{alg:adaptive_guidance_framework} presents the complete workflow of our method. The algorithm operates through seven interconnected phases:

\textbf{Phase 1: Experience Retrieval} encodes the input problem using TF-IDF and retrieves the top-3 most similar historical cases from the experience repository. These retrieved cases provide the weak agent with relevant solution patterns, reducing initial exploration cost.

\textbf{Phase 2: Response Generation}  allows the weak agent to produce an initial solution informed by both the problem and retrieved experiences. This response reflects the agent's current understanding level.

\textbf{Phase 3: Understanding Assessment}computes multi-dimensional entropy across five aspects: expression (vocabulary dispersion), uncertainty (semantic markers), structure (logical organization), coherence (reasoning consistency), and relevance (problem alignment). The aggregated entropy $H_u^{(t)}$ quantifies comprehension level—higher entropy indicates confusion, lower entropy reflects understanding.

\textbf{Phase 4: Adaptive Guidance Selection} chooses guidance intensity based on entropy thresholds. Low entropy ($H_u \leq \tau_1$) triggers light guidance with verification only. Moderate entropy ($\tau_1 < H_u \leq \tau_2$) provides conceptual frameworks and suggestions. High entropy ($H_u > \tau_2$) deploys intensive guidance with detailed problem decomposition. This tiered approach matches guidance to cognitive needs.

\textbf{Phase 5: Dynamic Threshold Adjustment} progressively reduces intervention as the weak agent improves. The adjustment amount $A(t)$ increases with rounds (learning rate $\lambda = 0.2$) but caps at $A_{\max} = 0.6$. Both thresholds decrease while respecting minimum bounds, gradually encouraging autonomy while maintaining support.

\textbf{Phase 6: Solution Verification} checks whether the weak agent's response correctly solves the problem, examining both the final answer and reasoning process.

\textbf{Phase 7: Experience Storage}creates structured records of successful collaborations, including problem text, solution steps, answer, difficulty level, success entropy, and usage count. These records are added to the repository for future retrieval. Unsuccessful attempts are not stored.

The iterative loop continues until a correct solution is found or the maximum round limit is reached. Each iteration refines understanding through tailored guidance, with entropy assessment providing continuous feedback. This combination of adaptive guidance, dynamic thresholds, and experience accumulation creates a virtuous cycle where weak agents not only solve problems effectively but also build long-term capabilities.

\begin{algorithm}[t]
\caption{Entropy-Based Adaptive Guidance for Heterogeneous Multi-Agent Collaboration}
\label{alg:adaptive_guidance_framework}
\begin{algorithmic}[1]
\REQUIRE Problem $P$, Task type $\mathcal{T}$, Strong agent $A_s$, Weak agent $A_w$, Experience repository $\mathcal{E}$
\ENSURE Final solution $S_{\text{final}}$, Updated repository $\mathcal{E}'$

\STATE \textbf{Initialize:} Round $t \leftarrow 0$, Thresholds $\tau_1 \leftarrow \tau_{1,\text{base}}, \tau_2 \leftarrow \tau_{2,\text{base}}$

\STATE Encode problem: $\mathbf{v}_P \leftarrow \text{TF-IDF}(P)$ \hfill
\STATE Retrieve: $\mathcal{E}_{\text{relevant}} \leftarrow \text{TopK-Similar}(\mathbf{v}_P, \mathcal{E}, k=3)$

\REPEAT
    \STATE $t \leftarrow t + 1$
    
    \STATE $R_w^{(t)} \leftarrow A_w.\text{Generate}(P, \mathcal{E}_{\text{relevant}})$
    
    \STATE Compute entropy components: $H_{\text{expr}}, H_{\text{unc}}, H_{\text{struct}}, H_{\text{coh}}, H_{\text{rel}}$ \hfill 
    \STATE Aggregate: $H_u^{(t)} \leftarrow H_{\text{expr}} + H_{\text{unc}} + H_{\text{struct}} - H_{\text{coh}} - H_{\text{rel}}$
    
    \IF{$H_u^{(t)} \leq \tau_1$}
        \STATE $G^{(t)} \leftarrow G_{\text{light}}(P, R_w^{(t)}, \mathcal{T})$ \hfill \textit{// Light: Verification + Correction}
    \ELSIF{$\tau_1 < H_u^{(t)} \leq \tau_2$}
        \STATE $G^{(t)} \leftarrow G_{\text{moderate}}(P, R_w^{(t)}, \mathcal{T})$ \hfill \textit{// Moderate: Framework + Suggestions}
    \ELSE
        \STATE $G^{(t)} \leftarrow G_{\text{intensive}}(P, R_w^{(t)}, \mathcal{T})$ \hfill \textit{// Intensive: Step-by-step Decomposition}
    \ENDIF
    \STATE Provide guidance: $A_s \rightarrow A_w: G^{(t)}$
    
    \STATE $A(t) \leftarrow \min(\lambda \cdot (t-1), A_{\max})$ \hfill 
    \STATE $\tau_1 \leftarrow \max(\tau_{1,\text{base}} - A(t), \tau_{1,\min})$ \hfill 
    \STATE $\tau_2 \leftarrow \max(\tau_{2,\text{base}} - A(t), \tau_{2,\min})$
    
    \STATE $\text{isCorrect} \leftarrow A_s.\text{Verify}(R_w^{(t)}, P)$
    
\UNTIL{$\text{isCorrect}$ \textbf{or} $t \geq t_{\max}$}

\STATE $S_{\text{final}} \leftarrow R_w^{(t)}$

\IF{$\text{isCorrect}$}
    \STATE $E_{\text{new}} \leftarrow \langle P, S_{\text{steps}}, A_{\text{final}}, D_{\text{level}}, H_u^{(t)}, 0 \rangle$ \hfill 
    \STATE $\mathcal{E}' \leftarrow \mathcal{E} \cup \{E_{\text{new}}\}$
\ELSE
    \STATE $\mathcal{E}' \leftarrow \mathcal{E}$
\ENDIF

\RETURN $S_{\text{final}}, \mathcal{E}'$
\end{algorithmic}
\end{algorithm}
\section{Experiments}
\subsection{Experimental setup}
\subsubsection{Datasets}We selected three benchmark datasets to evaluate the performance of the heterogeneous agent system across different problem-solving domains.

GSM8K~\cite{zeng2023mr}: This dataset focuses on mathematical reasoning tasks. We randomly sampled 50 problems and performed stratified sampling based on difficulty levels (Easy: 30\%, Medium: 50\%, Hard: 20\%) to ensure balanced task complexity.

MBPP~\cite{yu2025humaneval}: The Mostly Basic Python Problems dataset contains programming and algorithmic reasoning tasks. We randomly selected 50 problems to evaluate the agents’ ability in code generation and program understanding.

CVRP~\cite{ralphs2003capacitated}: The Capacitated Vehicle Routing Problem dataset represents combinatorial optimization scenarios. We randomly chose 20 tasks to test the system’s capability in multi-constraint decision-making and cooperative route planning.
\subsubsection{Models}We adopted two categories of large language models to represent strong and weak agents.
The strong agents include GPT-4o~\cite{pang2024chatgpt} and Claude-3.5-Sonnet~\cite{bae2024enhancing}, while the weak agents include Qwen-2.5-0.5B~\cite{hui2024qwen2} and Llama-3.2-1B~\cite{grattafiori2024llama}.
In each experiment, we formed a strong–weak pair by selecting one model from each category to simulate heterogeneous cooperation.
\subsubsection{Collaboration Strategies}Different collaboration schemes were designed according to the task characteristics of each dataset.

GSM8K: A Framework–Solver collaboration is used. The Framework Agent generates structured reasoning outlines, while the Solver Agent fills in detailed logical steps and derives the final answer, forming a process from structured reasoning to precise computation.
MBPP: A Framework Provider + Implementer mode is adopted. The Framework Provider produces high-level program structures and annotated pseudocode, while the Implementer completes the concrete implementation and debugging, improving both correctness and readability of the generated code.

CVRP: A Proposer–Validator collaboration is employed. The Proposer Agent generates candidate routing solutions, and the Validator Agent evaluates them based on task constraints and provides optimization feedback. Through multiple rounds of interaction, the two agents iteratively improve the feasibility and optimality of the solutions.
\subsubsection{Evaluation Metrics}For GSM8K, we use Accuracy, measuring the proportion of correctly solved mathematical reasoning problems. For MBPP, we adopt Final Pass Rate as the main metric, representing the proportion of correct code generations after guidance or multiple interaction rounds. We also compute the Improvement Rate, indicating the proportion of problems that changed from incorrect to correct during the guidance process. For CVRP, we use Mean Accuracy, which measures the average proximity of generated routes to the optimal solution, averaged across all successful trials to reflect overall optimization performance.

\subsubsection{Baselines}
No-Guidance Baseline: In this setting, a single agent completes each task independently without any external guidance or collaboration. The results reflect the model’s inherent reasoning ability and serve as the reference for evaluating the impact of cooperative mechanisms.
Chain-of-Thought (CoT) Baseline: In this baseline, a single agent is prompted with Chain-of-Thought reasoning cues to encourage step-by-step problem solving. This setting is used to assess the performance improvement achieved through conventional explicit reasoning enhancement compared to our collaborative framework.
\subsection{Experimental results}
\subsubsection{Performance Across Different Agent Combinations}

\begin{table*}[ht!]
\centering
\caption{Performance of Heterogeneous Agent Combinations across Benchmark Datasets}
\label{tab:agent_combinations_simple}
\begin{tabular}{l@{\hspace{0.5cm}}l@{\hspace{0.5cm}}l@{\hspace{1cm}}|@{\hspace{1cm}}c@{\hspace{1cm}}c@{\hspace{1cm}}c}
\toprule
\multicolumn{3}{c|}{\textbf{Agent Configuration}} & \multicolumn{3}{c}{\textbf{Performance (\%)}} \\
\cmidrule(lr){1-3} \cmidrule(lr){4-6}
\textbf{Type} & \textbf{Guide} & \textbf{Solver} & 
  \textbf{GSM8K} & \textbf{MBPP} & \textbf{CVRP} \\
\midrule
\multirow{2}{*}{\textbf{Strong-Strong}} 
  & GPT-4o & GPT-4o & \textbf{74.0} & \textbf{74.0} & \textbf{78.5} \\
  & Claude-3.5-Sonnet & Claude-3.5-Sonnet & \textbf{84.0} & \textbf{84.0} & \textbf{87.5} \\
\midrule
\multirow{4}{*}{\textbf{Strong-Weak}} 
  & GPT-4o & Qwen-2.5-0.5B & 46.0 & 20.0 & 66.8 \\
  & Claude-3.5-Sonnet & Qwen-2.5-0.5B & 48.0 & 22.0 & 64.8 \\
  & GPT-4o & Llama-3.2-1B & 48.0 & 50.0 & 62.8 \\
  & Claude-3.5-Sonnet & Llama-3.2-1B & 38.0 & 54.0 & 65.5 \\
\midrule
\multirow{2}{*}{\textbf{Weak-Weak}} 
  & Llama-3.2-1B & Llama-3.2-1B & \underline{46.0} & \underline{54.0} & \underline{69.2} \\
  & Qwen-2.5-0.5B & Qwen-2.5-0.5B & \underline{58.0} & \underline{26.0} & \underline{69.5} \\
\bottomrule
\end{tabular}
\caption*{\footnotesize \textbf{Bold}: Best performance. \underline{Underline}: Weak-weak baseline.}
\end{table*}

To systematically verify the impact of heterogeneous capability gaps on collaboration effectiveness, we compared the performance of three different strong-weak agent combinations across benchmark datasets. Table~\ref{tab:agent_combinations_simple} presents detailed experimental results.

Strong-strong collaborations achieved optimal performance across all tasks. Claude-3.5-Sonnet's self-collaboration configuration reached accuracies of 84.0\%, 84.0\%, and 87.5\% on GSM8K, MBPP, and CVRP respectively, significantly outperforming other configurations. GPT-4o's self-collaboration also demonstrated strong results with accuracies of 74.0\%, 74.0\%, and 78.5\% across the three tasks.

Performance declined significantly when strong agents GPT or Claude guided weak agents Qwen or Llama. For instance, GPT guiding Qwen achieved only 46.0\% accuracy on GSM8K, 28\% drop compared to GPT self-collaboration. More surprisingly, this pairing achieved merely 20.0\% accuracy on MBPP, falling below Qwen self-collaboration performance of 26.0\%.    

Weak-weak collaborations provided an important baseline for comparison. Llama's self-collaboration achieved 46.0\%, 54.0\%, and 69.2\% on the three tasks, while Qwen's self-collaboration reached 58.0\%, 26.0\%, and 69.5\%. Notably, strong-weak combinations failed to surpass these weak-weak baselines in multiple scenarios, further confirming the cognitive mismatch issues in heterogeneous collaboration.

Different tasks showed varying sensitivity to heterogeneous collaboration. On CVRP, strong-weak combinations experienced relatively modest performance drops mostly maintaining 60-66\%, while tasks requiring precise reasoning like GSM8K and MBPP showed more severe degradation. This indicates that cognitive complexity directly affects heterogeneous collaboration effectiveness.

These results clearly demonstrate that in heterogeneous multi-agent systems, simply pairing strong and weak agents fails to improve overall performance effectively. Instead, the substantial cognitive capability gap may lead to collaboration failure.

\subsubsection{Effectiveness of Entropy-Based Adaptive Guidance
}
\begin{table*}[ht!]
\centering
\caption{Performance Comparison Across Methods and Agent Pairs}
\label{tab:guidance_horizontal}
\begin{threeparttable}
\small
\begin{tabular}{l|cccc|c}
\toprule
& \multicolumn{4}{c|}{\textbf{Agent Pairs}} & \\
\cmidrule(lr){2-5}
\textbf{Method} & \textbf{GPT+Llama} & \textbf{Claude+Llama} & \textbf{GPT+Qwen} & \textbf{Claude+Qwen} & \textbf{Avg.} \\
\midrule
\multicolumn{6}{c}{\textit{GSM8K (Accuracy \%)}} \\
\midrule
No-Guidance & 48.0 & 38.0 & 46.0 & 48.0 & 45.0 \\
Chain-of-Thought & 56.0 & 46.0 & 48.0 & 54.0 & 51.0 \\
Guided (Ours) & 62.0 & 74.0 & 60.0 & 60.0 & 64.0 \\
\textbf{Guided+RAG (Ours)} & \textbf{62.0} & \textbf{78.0} & \textbf{68.0} & \textbf{68.0} & \textbf{69.0} \\
\midrule
\multicolumn{6}{c}{\textit{MBPP (Pass Rate \%)}} \\
\midrule
No-Guidance & 50.0 & 54.0 & 20.0 & 22.0 & 36.5 \\
Chain-of-Thought & 52.0 & 56.0 & 40.0 & 26.0 & 43.5 \\
Guided (Ours) & 52.0 & 54.0 & 44.0 & 24.0 & 43.5 \\
\textbf{Guided+RAG (Ours)} & \textbf{54.0} & \textbf{54.0} & \textbf{46.0} & \textbf{30.0} & \textbf{46.0} \\
\midrule
\multicolumn{6}{c}{\textit{CVRP (Accuracy \%)}} \\
\midrule
No-Guidance & 62.8 & 65.5 & 66.8 & 64.8 & 65.0 \\
Chain-of-Thought & 57.3 & 66.6 & 69.0 & 73.4 & 66.6 \\
Guided (Ours) & 76.9 & 63.8 & 73.1 & 72.6 & 71.6 \\
\textbf{Guided+RAG (Ours)} & \textbf{85.9} & \textbf{77.7} & \textbf{87.2} & \textbf{74.5} & \textbf{81.3} \\
\bottomrule
\end{tabular}
\begin{tablenotes}
\footnotesize
\item GPT = GPT-4o, Claude = Claude-3.5-Sonnet, Llama = Llama-3.2-1B, Qwen = Qwen-2.5-0.5B.
\item \textbf{Bold}: Best performance with RAG enhancement.
\item Values averaged across all test instances in each dataset.
\end{tablenotes}
\end{threeparttable}
\end{table*}
% \begin{figure*}[ht!]
%     \centering
%     \includegraphics[width=0.95\textwidth]{fig/performance.pdf}
%     \caption{Performance gains of our entropy-based adaptive guidance methods compared to No-Guidance and Chain-of-Thought baselines across three benchmark datasets. }
%     \label{fig:performance_gains}
% \end{figure*}

To validate the effectiveness of our entropy-based adaptive guidance approach, we designed comprehensive comparative experiments comparing our method against two baseline approaches: (1) No-Guidance, where weak agents work independently; and (2) Chain-of-Thought (CoT) prompting, representing conventional non-adaptive guidance. All methods were evaluated on the same strong-weak agent pairs identified previously. 

Table~\ref{tab:guidance_horizontal} presents the performance comparison across different methods and agent pairs, demonstrating consistent improvements of our adaptive guidance approach across all three benchmark tasks.
On the GSM8K mathematical reasoning task, our guided approach achieved 64.0\% average accuracy, representing a 19.0 percentage point improvement over the No-Guidance baseline (45.0\%) and a 13.0 point gain over CoT (51.0\%). With RAG enhancement, performance further increased to 69.0\%, achieving a remarkable 24.0 percentage point improvement over the No-Guidance baseline. 

Notably, the Claude and Llama pair showed the most significant improvement, rising from 38.0\% to 78.0\%.
For the MBPP code generation task, while improvements were more modest, our method still demonstrated effectiveness. The guided approach matched CoT performance (43.5\%), while Guided and RAG reached 46.0\%, a 9.5 percentage point improvement over No-Guidance. The GPT+Qwen pair showed the largest gains, improving from 20.0\% to 46.0\%.

On the CVRP combinatorial optimization task, our method showed the strongest response. The guided approach achieved 71.6\% accuracy, with Guided+RAG reaching 81.3\%—a 16.3 percentage point improvement over No-Guidance and 14.7 points over CoT. The GPT+Llama pair performed exceptionally well, achieving 85.9\% accuracy with RAG enhancement.
\begin{figure}[t]
    \centering
    \includegraphics[width=\columnwidth]{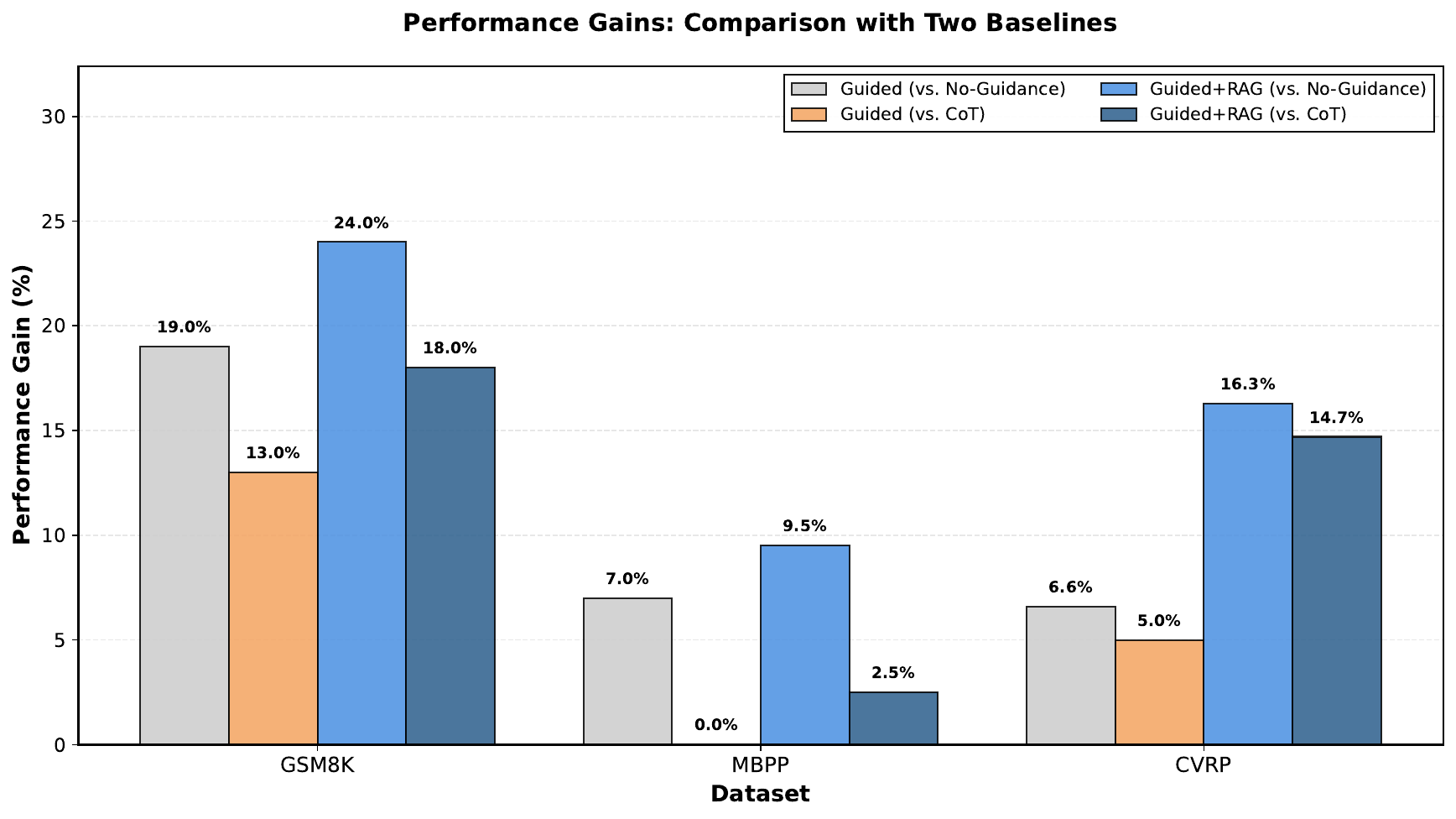}
    \caption{Performance gains of our entropy-based adaptive guidance methods compared to No-Guidance and Chain-of-Thought baselines across three benchmark datasets.}
    \label{fig:performance_gains}
\end{figure}

Figure~\ref{fig:performance_gains} illustrates the relative performance gains of our method compared to both baselines across all datasets. Several key patterns emerge from these results. Our guided approach consistently outperformed both baselines across all tasks, with gains ranging from 6.6\% to 24.0\%, validating the robustness of our entropy-based adaptive strategy. The addition of experience retrieval provided substantial additional gains, particularly on GSM8K (5.0\% additional improvement) and CVRP (9.7\% additional improvement), demonstrating that leveraging historical successful experiences effectively complements real-time guidance.

Mathematical reasoning and combinatorial optimization  showed stronger responses to our approach compared to code generation , consistent with our hypothesis that tasks requiring structured reasoning benefit more from adaptive guidance. While CoT provided some improvement over No-Guidance, our adaptive approach significantly outperformed it, demonstrating that static guidance strategies are insufficient for addressing cognitive gaps in heterogeneous collaboration.

These results confirm that our entropy-based adaptive guidance successfully addresses the performance bottlenecks identified in Section~\ref{sec:4.1}, enabling weak agents to better utilize knowledge provided by strong agents through calibrated, task-appropriate guidance levels.

\subsection{Ablation Study}
\subsubsection{Performance Under Different Collaboration Strategies}

\begin{figure}[t]
    \centering
    \includegraphics[width=0.48\textwidth]{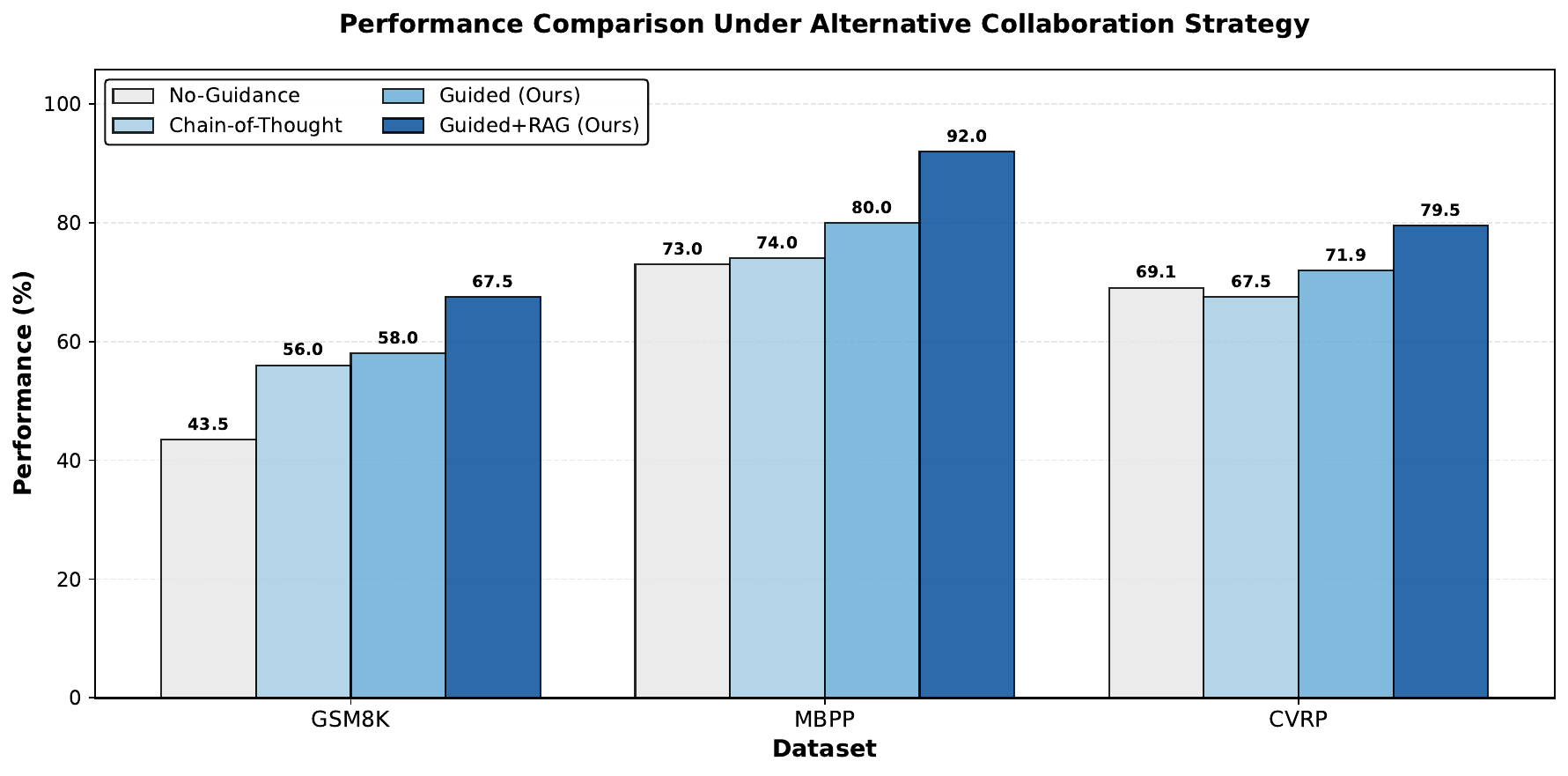}
    \caption{Comparative performance of four collaboration strategies (No-Guidance, Chain-of-Thought, Guided, and Guided+RAG) across three benchmark datasets. Results demonstrate consistent superiority of our entropy-based adaptive guidance approach, with Guided+RAG achieving the highest performance across all tasks.}
    \label{fig:strategy_methods}
\end{figure}

To further validate the generalizability of our approach, this section analyzes the performance of different collaboration strategies across various agent combinations. Notably, the three benchmark tasks employ different collaboration modes: GSM8K uses a "solve-review" pattern, MBPP adopts a "code-check-improve" pattern, and CVRP follows a "sequential collaboration" pattern, strong agent initiates and weak agent completes. These diverse collaboration modes provide ideal test scenarios for evaluating the generalization capability of our method.

As shown in Figure~\ref{fig:strategy_methods}, we observe a clear performance gradient across the four collaboration strategies on three benchmark tasks. In GSM8K's solve-review mode, where weak agents first attempt to solve problems followed by strong agent review and correction, the Guided+RAG strategy achieved 67.5\% accuracy, a 24 percentage point improvement over the No-Guidance baseline (43.5\%). This demonstrates that our adaptive guidance mechanism effectively enhances the initial problem-solving quality of weak agents, reducing the correction burden on strong agents.

Under MBPP's code-check-improve mode, the task showed the most significant improvements. The Guided+RAG strategy achieved an exceptional 92.0\% performance, far exceeding No-Guidance (73.0\%) and CoT (74.0\%). In this collaboration mode where one agent writes code and another checks and improves it, our method significantly enhanced both code quality and collaboration efficiency.

In CVRP's sequential collaboration mode, where strong agents initiate and weak agents complete tasks, performance improvements were relatively modest but still significant. Guided+RAG reached 79.5\%, a 10.4 percentage point improvement over the baseline method (69.1\%). Under this mode, adaptive guidance helped weak agents better understand and execute the planning solutions provided by strong agents.

% \begin{figure}[t]
%     \centering
%     \includegraphics[width=0.48\textwidth]{fig/strategy_b_boxplot.pdf}
%     \caption{Performance distribution of different agent configurations (Strong-Strong, Strong-Weak, Weak-Weak) across three benchmark datasets under various collaboration strategies. Box plots show median, quartiles, and outliers, revealing how our adaptive guidance methods reduce performance variance and improve median performance, particularly for Strong-Weak combinations.}
%     \label{fig:strategy_boxplot}
% \end{figure}

\begin{figure*}[t]
    \centering
    \includegraphics[width=0.95\textwidth]{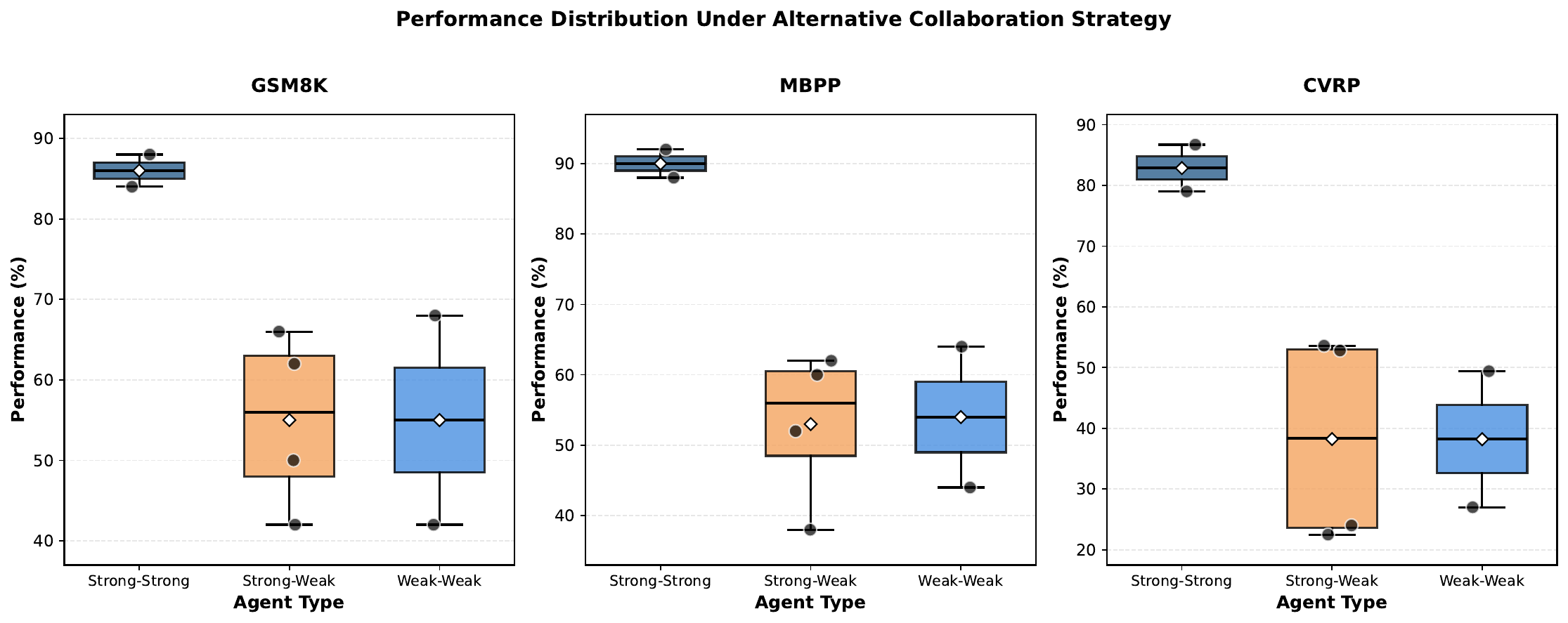}
    \caption{Performance distribution of different agent configurations (Strong-Strong, Strong-Weak, Weak-Weak) across three benchmark datasets under various collaboration strategies. Box plots show median, quartiles, and outliers, revealing how our adaptive guidance methods reduce performance variance and improve median performance, particularly for Strong-Weak combinations.}
    \label{fig:strategy_boxplot}
\end{figure*}

Figure~\ref{fig:strategy_boxplot} reveal performance distribution characteristics of different collaboration strategies across three agent configurations. Across all collaboration modes, strong-strong combinations consistently exhibited high performance with low variance. Whether in GSM8K's solve-review mode (median 86\%), MBPP's code-check mode (approaching 90\%), or CVRP's sequential collaboration (exceeding 83\%), collaborations between strong agents maintained stable high-level performance.

In GSM8K's solve-review mode, performance distributions improved significantly from the low levels under No-Guidance (median about 55\%), indicating substantial enhancement in weak agents' problem-solving capabilities. MBPP's code-check mode showed the largest improvement magnitude with significantly reduced box heights, demonstrating that our method effectively improved weak agents' code writing quality and reduced performance fluctuations. While CVRP's sequential collaboration mode exhibited a larger performance span (from about 25\% to 55\%), the median improvement was notable, indicating enhanced understanding and execution capabilities of weak agents when implementing strong agents' plans.

Weak-weak combinations showed differentiated characteristics across collaboration modes. They maintained moderate levels on reasoning-intensive GSM8K and MBPP tasks (medians 55\% and 53\% respectively), while performing lower on planning-execution CVRP (median 40\%), reflecting capability differences of weak agents across different cognitive tasks.

Regardless of collaboration mode, our adaptive guidance strategy produced maximum benefits for strong-weak combinations, successfully narrowing the performance gap with strong-strong combinations and validating the method's effectiveness in addressing cognitive mismatches in heterogeneous collaboration. Beyond average performance improvements, our method reduced performance variance of strong-weak combinations across all collaboration modes, demonstrating that adaptive guidance not only enhanced collaboration effectiveness but also improved reliability across different collaboration patterns.

\subsubsection{Impact of Experience Retrieval (RAG) Mechanism}

To evaluate the contribution of the RAG-based experience retention mechanism, we conducted a detailed ablation study on the CVRP task. We compared the performance of all four agent pairs with and without RAG enhancement. Table~\ref{tab:rag} presents comprehensive metrics including accuracy, improvement rate, solution quality, and RAG usage frequency.

\begin{table*}[t]
\centering
\caption{Detailed RAG Ablation Study across Agent Pairs on CVRP}
\label{tab:rag}
\begin{tabular}{l|cc|cc|cc|c}
\toprule
\multirow{2}{*}{\textbf{Agent Pair}} & \multicolumn{2}{c|}{\textbf{Accuracy (\%)}} & \multicolumn{2}{c|}{\textbf{Improvement}} & \multicolumn{2}{c|}{\textbf{Quality}} & \multirow{2}{*}{\textbf{RAG Usage}} \\
\cmidrule(lr){2-3} \cmidrule(lr){4-5} \cmidrule(lr){6-7}
& Guided & +RAG & Guided & +RAG & Guided & +RAG & \\
\midrule
GPT + Llama & 76.9 & 85.9 & 1.99 & 1.57 & 5.97 & 7.25 & 5\% \\
Claude + Llama & 63.8 & 77.7 & -0.77 & 8.99 & 6.63 & 6.83 & 41.6\% \\
GPT + Qwen & 73.1 & 87.2 & 0.32 & 4.4 & 5.71 & 6.87 & 3.33\% \\
Claude + Qwen & 72.6 & 74.5 & 0.2 & 12.03 & 8.31 & 7.02 & 38.3\% \\
\midrule
\textbf{Average} & 71.6 & 81.3 & 0.08 & 0.12 & 6.4 & 7.1 & 38.3\% \\
\bottomrule
\end{tabular}
\end{table*}

The addition of RAG consistently improved collaboration effectiveness across all agent pairs. On average, RAG increased accuracy from 71.6\% to 81.3\%, achieving a significant gain of 9.7 percentage points. Different agent pairs showed different levels of improvement: GPT+Qwen achieved the largest accuracy gain (+14.1 percentage points, from 73.1\% to 87.2\%), followed by GPT+Llama (+9.0 percentage points) and Claude+Llama (+13.9 percentage points). Even the smallest improvement (Claude+Qwen, +1.9 percentage points) demonstrates the positive effect of RAG.

Beyond accuracy metrics, RAG also significantly improved solution quality scores. The average quality increased from 6.4 to 7.1, indicating that RAG not only helps weak agents reach correct solutions more frequently but also enhances the overall optimization quality of generated routes. Notably, GPT+Llama showed the most significant quality improvement (from 5.97 to 7.25), while Claude+Qwen exhibited high baseline quality (8.31) and maintained strong performance with RAG (7.02).

The improvement rate metric reveals interesting patterns in collaboration efficiency. Without RAG, the average improvement rate was close to zero (0.08), suggesting that many solutions required extensive iterations to converge. With RAG, this metric increased slightly to 0.12, but more importantly, its distribution changed significantly. Claude+Qwen showed a remarkable improvement rate of 12.03 with RAG, indicating that experience retrieval enabled this pairing to converge to optimal solutions much faster. Claude+Llama's negative improvement rate without RAG (-0.77) transformed into a strong positive value (8.99), fully demonstrating RAG's effectiveness in stabilizing challenging collaborations.

RAG usage frequency varied significantly across different agent pairs, revealing adaptive retrieval behavior. Claude-based pairs showed notably higher retrieval rates (Claude+Llama: 41.6\%, Claude+Qwen: 38.3\%), while GPT-based pairs had lower retrieval rates (GPT+Llama: 5\%, GPT+Qwen: 3.33\%). This suggests that when Claude serves as the guide, weak agents face greater initial uncertainty and therefore rely more heavily on historical experiences. In contrast, GPT's guidance may be more immediately understandable, reducing the need for experience retrieval. Despite lower usage frequency, GPT-based pairs still achieved competitive or superior accuracy gains, indicating that when retrieval occurs, it is highly targeted and effective.

Different weak agents showed distinct response patterns to RAG enhancement. Qwen benefited significantly when paired with GPT (14.1 percentage point gain) but showed minimal improvement with Claude (1.9 percentage points), despite Claude+Qwen's high RAG usage rate (38.3\%). This suggests that the effectiveness of experience retrieval depends not only on usage frequency but also on the compatibility between guidance style and agent cognitive characteristics. In contrast, Llama showed more consistent gains across both strong agents (9.0 and 13.9 percentage point gains), indicating its ability to learn from historical experiences more robustly.

The non-linear relationship between RAG usage frequency and performance improvement reveals the underlying mechanism of experience retrieval. The high usage rate in Claude pairs (approximately 40\%) reflects the higher cognitive load faced by weak agents in these collaboration scenarios—Claude as a guide may provide more abstract or conceptual suggestions, prompting weak agents to seek concrete historical cases as references more frequently. The low usage rate but high effectiveness in GPT pairs (such as GPT+Qwen with only 3.33\% usage rate yet achieving 14.1\% accuracy improvement) suggests that GPT's guidance style may be more concrete and structured, allowing weak agents to need experience support only at critical decision points, but each retrieval precisely addresses core confusion.

These ablation results validate our hypothesis: experience retention through RAG significantly enhances the effectiveness of heterogeneous collaboration. The mechanism provides dual benefits: (1) reducing initial understanding entropy by providing relevant precedents, and (2) enabling weak agents to leverage accumulated successful patterns, thereby accelerating convergence and improving solution quality. The diverse usage patterns across different agent pairs further confirm RAG's adaptive operation—when collaboration faces greater cognitive challenges, the system invokes the experience retrieval mechanism more frequently.

\subsection{Communication Efficiency Analysis}

% \begin{figure*}[t]
%     \centering
%     \includegraphics[width=0.95\textwidth]{fig/strategy_b_boxplot.pdf}
%     \caption{Performance distribution of different agent configurations (Strong-Strong, Strong-Weak, Weak-Weak) across three benchmark datasets under various collaboration strategies. Box plots show median, quartiles, and outliers, revealing how our adaptive guidance methods reduce performance variance and improve median performance, particularly for Strong-Weak combinations.}
%     \label{fig:strategy_boxplot}
% \end{figure*}

% \begin{figure*}[t]
%     \centering
%     \includegraphics[width=0.95\textwidth]{fig/strategy_b_methods.pdf}
%     \caption{Comparative performance of four collaboration strategies (No-Guidance, Chain-of-Thought, Guided, and Guided+RAG) across three benchmark datasets. Results demonstrate consistent superiority of our entropy-based adaptive guidance approach, with Guided+RAG achieving the highest performance across all tasks.}
%     \label{fig:strategy_methods}
% \end{figure*}

\begin{figure*}[t]
    \centering
    \includegraphics[width=0.95\textwidth]{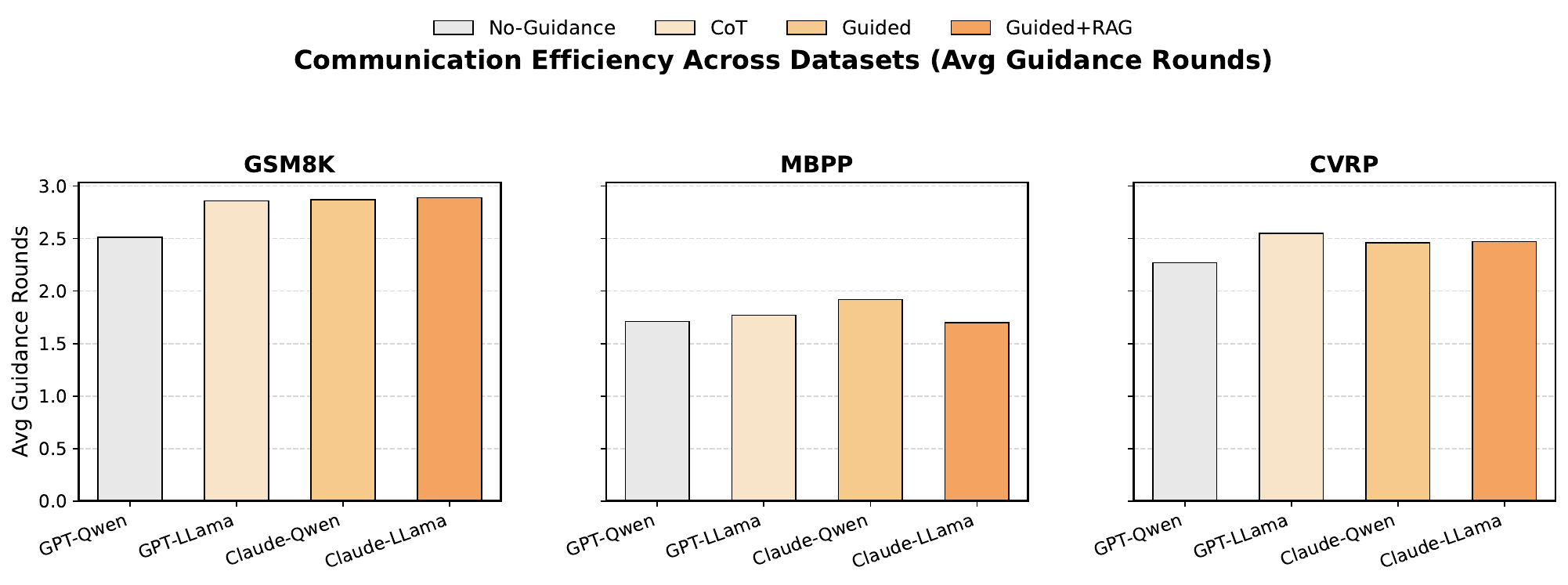}
    \caption{Communication efficiency analysis across different agent pairs and benchmark tasks. The figure shows average guidance rounds required for each agent configuration, revealing task-specific communication patterns where GSM8K requires the most rounds (2.5-2.9), CVRP shows moderate requirements (2.3-2.6), and MBPP demonstrates the highest efficiency (1.7-1.9 rounds).}
    \label{fig:communication_efficiency}
\end{figure*}
This section quantifies the communication overhead of different collaboration strategies by analyzing average guidance rounds, revealing how adaptive guidance impacts communication efficiency while enhancing performance. 

As shown in Figure~\ref{fig:communication_efficiency}, we present the average guidance rounds for four collaboration strategies across different agent pairs and tasks. Overall, our adaptive guidance method demonstrates good communication efficiency while maintaining high performance.

GSM8K reasoning tasks exhibited the highest communication demands with average rounds ranging from 2.51 to 2.89. MBPP code generation tasks showed the lowest communication requirements, with all pairs requiring only 1.7-1.92 rounds. CVRP combinatorial optimization tasks fell between these extremes, requiring 2.27-2.55 rounds. This gradient clearly reflects the inherent complexity of different cognitive tasks—mathematical reasoning requires the most interaction to ensure conceptual understanding, while the structured nature of code generation necessitates minimal guidance.

On MBPP tasks, both strong agents performed comparably as guides. On CVRP tasks, GPT showed a slight advantage as a guide. On GSM8K tasks, the GPT-Qwen combination significantly outperformed other pairs, potentially reflecting specific compatibility advantages.

As weak agent receivers, Qwen and LLama exhibited different adaptability across tasks. On MBPP, LLama slightly outperformed Qwen, while on CVRP, their performance was similar. The superior performance of GPT-Qwen on GSM8K suggests certain pairs may possess special cognitive compatibility. Claude-LLama's efficiency on MBPP likely stems from Claude's clear code explanation style matching well with LLama's comprehension capabilities.

These results indicate that while adaptive guidance strategies require additional communication rounds, this overhead is reasonable and efficient. Through precise entropy assessment and tiered guidance, the system maintains performance improvements while keeping communication costs within acceptable bounds, providing strong efficiency guarantees for practical applications.

\section{Conclusion}
This paper addresses a critical yet underexplored challenge in heterogeneous multi-agent systems: the cognitive bottleneck arising from capability asymmetry between strong and weak agents. Through systematic experimentation, we uncover a counterintuitive negative synergy effect—in certain tasks, strong-weak collaboration performs even worse than weak-weak collaboration. Our analysis reveals that this degradation stems from cognitive overload and inefficient information transfer when weak agents struggle to interpret complex guidance from stronger counterparts.

To mitigate this issue, we propose the Entropy-Based Adaptive Guidance Framework, which redefines how heterogeneous agents collaborate. By quantifying weak agents’ understanding states via multi-dimensional entropy metrics, the framework enables strong agents to deliver guidance dynamically aligned with weak agents’ cognitive capacities. It incorporates three core innovations: (1) an entropy-based understanding assessment mechanism capturing fine-grained cognitive states across five dimensions, (2) a three-level adaptive guidance strategy with dynamic threshold adjustment that calibrates intervention intensity over time, and (3) a RAG-based experience retention module that facilitates continual learning through structured reuse of past successes.

Extensive experiments on GSM8K, MBPP, and CVRP benchmarks demonstrate the effectiveness of our framework, yielding substantial performance gains of 9.5–24.0 percentage points over baseline methods. Beyond technical contributions, our findings provide broader insights into heterogeneous collaboration: (1) performance is ultimately constrained by the weakest agent’s ability to assimilate information, rather than the strongest agent’s capability; (2) adaptive, capability-aware guidance is essential for effective cooperation; and (3) integrating real-time guidance with long-term experience accumulation fosters sustainable improvement in multi-agent systems.
% \section*{Acknowledgments}
% This should be a simple paragraph before the References to thank those individuals and institutions who have supported your work on this article.

% {\appendix[Proof of the Zonklar Equations]
% Use $\backslash${\tt{appendix}} if you have a single appendix:
% Do not use $\backslash${\tt{section}} anymore after $\backslash${\tt{appendix}}, only $\backslash${\tt{section*}}.
% If you have multiple appendixes use $\backslash${\tt{appendices}} then use $\backslash${\tt{section}} to start each appendix.
% You must declare a $\backslash${\tt{section}} before using any $\backslash${\tt{subsection}} or using $\backslash${\tt{label}} ($\backslash${\tt{appendices}} by itself
%  starts a section numbered zero.)}


% \section{References Section}
% You can use a bibliography generated by BibTeX as a .bbl file.
 % The IEEEtran BibTeX style support page is:
 

\bibliographystyle{IEEEtran}
\bibliography{ref}

@article{ye2025x,
  title={X-MAS: Towards Building Multi-Agent Systems with Heterogeneous LLMs},
  author={Ye, Rui and Liu, Xiangrui and Wu, Qimin and Pang, Xianghe and Yin, Zhenfei and Bai, Lei and Chen, Siheng},
  journal={arXiv preprint arXiv:2505.16997},
  year={2025}
}

@article{yang2025autohma,
  title={AutoHMA-LLM: Efficient task coordination and execution in heterogeneous multi-agent systems using hybrid large language models},
  author={Yang, Tingting and Feng, Ping and Guo, Qixin and Zhang, Jindi and Ning, Jiahong and Wang, Xinghan and Mao, Zhongyang},
  journal={IEEE Transactions on Cognitive Communications and Networking},
  year={2025},
  publisher={IEEE}
}

@inproceedings{wu2024autogen,
  title={Autogen: Enabling next-gen LLM applications via multi-agent conversations},
  author={Wu, Qingyun and Bansal, Gagan and Zhang, Jieyu and Wu, Yiran and Li, Beibin and Zhu, Erkang and Jiang, Li and Zhang, Xiaoyun and Zhang, Shaokun and Liu, Jiale and others},
  booktitle={First Conference on Language Modeling},
  year={2024}
}

@article{gao2402agentscope,
  title={Agentscope: A flexible yet robust multi-agent platform, 2024},
  author={Gao, Dawei and Li, Zitao and Pan, Xuchen and Kuang, Weirui and Ma, Zhijian and Qian, Bingchen and Wei, Fei and Zhang, Wenhao and Xie, Yuexiang and Chen, Daoyuan and others},
  journal={URL https://arxiv. org/abs/2402.14034}
}

@article{wang2024megaagent,
  title={Megaagent: A practical framework for autonomous cooperation in large-scale llm agent systems},
  author={Wang, Qian and Wang, Tianyu and Li, Qinbin and Liang, Jingsheng and He, Bingsheng},
  journal={arXiv e-prints},
  pages={arXiv--2408},
  year={2024}
}

@article{ning2023skeleton,
  title={Skeleton-of-thought: Prompting llms for efficient parallel generation},
  author={Ning, Xuefei and Lin, Zinan and Zhou, Zixuan and Wang, Zifu and Yang, Huazhong and Wang, Yu},
  journal={arXiv preprint arXiv:2307.15337},
  year={2023}
}

@article{qiao2024autoact,
  title={Autoact: Automatic agent learning from scratch for qa via self-planning},
  author={Qiao, Shuofei and Zhang, Ningyu and Fang, Runnan and Luo, Yujie and Zhou, Wangchunshu and Jiang, Yuchen Eleanor and Lv, Chengfei and Chen, Huajun},
  journal={arXiv preprint arXiv:2401.05268},
  year={2024}
}

@article{suzgun2024meta,
  title={Meta-prompting: Enhancing language models with task-agnostic scaffolding},
  author={Suzgun, Mirac and Kalai, Adam Tauman},
  journal={arXiv preprint arXiv:2401.12954},
  year={2024}
}

@article{yang2025agentnet,
  title={Agentnet: Decentralized evolutionary coordination for llm-based multi-agent systems},
  author={Yang, Yingxuan and Chai, Huacan and Shao, Shuai and Song, Yuanyi and Qi, Siyuan and Rui, Renting and Zhang, Weinan},
  journal={arXiv preprint arXiv:2504.00587},
  year={2025}
}

@inproceedings{jeyakumar2024advancing,
  title={Advancing agentic systems: Dynamic task decomposition, tool integration and evaluation using novel metrics and dataset},
  author={Jeyakumar, Shankar Kumar and Ahmad, Alaa Alameer and Gabriel, Adrian Garret},
  booktitle={NeurIPS 2024 Workshop on Open-World Agents},
  year={2024}
}

@article{liang2023encouraging,
  title={Encouraging divergent thinking in large language models through multi-agent debate},
  author={Liang, Tian and He, Zhiwei and Jiao, Wenxiang and Wang, Xing and Wang, Yan and Wang, Rui and Yang, Yujiu and Shi, Shuming and Tu, Zhaopeng},
  journal={arXiv preprint arXiv:2305.19118},
  year={2023}
}

@article{chen2023autoagents,
  title={Autoagents: A framework for automatic agent generation},
  author={Chen, Guangyao and Dong, Siwei and Shu, Yu and Zhang, Ge and Sesay, Jaward and Karlsson, B{\"o}rje F and Fu, Jie and Shi, Yemin},
  journal={arXiv preprint arXiv:2309.17288},
  year={2023}
}

@article{li2023camel,
  title={Camel: Communicative agents for" mind" exploration of large language model society},
  author={Li, Guohao and Hammoud, Hasan and Itani, Hani and Khizbullin, Dmitrii and Ghanem, Bernard},
  journal={Advances in Neural Information Processing Systems},
  volume={36},
  pages={51991--52008},
  year={2023}
}

@inproceedings{liu2024dynamic,
  title={A dynamic llm-powered agent network for task-oriented agent collaboration},
  author={Liu, Zijun and Zhang, Yanzhe and Li, Peng and Liu, Yang and Yang, Diyi},
  booktitle={First Conference on Language Modeling},
  year={2024}
}

@article{chan2023chateval,
  title={Chateval: Towards better llm-based evaluators through multi-agent debate},
  author={Chan, Chi-Min and Chen, Weize and Su, Yusheng and Yu, Jianxuan and Xue, Wei and Zhang, Shanghang and Fu, Jie and Liu, Zhiyuan},
  journal={arXiv preprint arXiv:2308.07201},
  year={2023}
}

@article{wei2022chain,
  title={Chain-of-thought prompting elicits reasoning in large language models},
  author={Wei, Jason and Wang, Xuezhi and Schuurmans, Dale and Bosma, Maarten and Xia, Fei and Chi, Ed and Le, Quoc V and Zhou, Denny and others},
  journal={Advances in neural information processing systems},
  volume={35},
  pages={24824--24837},
  year={2022}
}

@article{sanwal2025layered,
  title={Layered chain-of-thought prompting for multi-agent llm systems: A comprehensive approach to explainable large language models},
  author={Sanwal, Manish},
  journal={arXiv preprint arXiv:2501.18645},
  year={2025}
}

@article{sani2024towards,
  title={Towards More Explainable AI: Layered Chain-of-Thought Prompting in Multi-Agent Systems},
  author={Sani, Awais and Wahab, Abdul},
  year={2024}
}

@article{low2025surgraw,
  title={Surgraw: Multi-agent workflow with chain-of-thought reasoning for surgical intelligence},
  author={Low, Chang Han and Wang, Ziyue and Zhang, Tianyi and Zeng, Zhitao and Zhuo, Zhu and Mazomenos, Evangelos B and Jin, Yueming},
  journal={arXiv preprint arXiv:2503.10265},
  year={2025}
}

@inproceedings{du2023improving,
  title={Improving factuality and reasoning in language models through multiagent debate},
  author={Du, Yilun and Li, Shuang and Torralba, Antonio and Tenenbaum, Joshua B and Mordatch, Igor},
  booktitle={Forty-first International Conference on Machine Learning},
  year={2023}
}

@article{zhang2025if,
  title={If multi-agent debate is the answer, what is the question},
  author={Zhang, Hangfan and Cui, Zhiyao and Wang, Xinrun and Zhang, Qiaosheng and Wang, Zhen and Wu, Dinghao and Hu, Shuyue},
  journal={arXiv preprint arXiv:2502.08788},
  year={2025}
}

@article{chen2024magicore,
  title={Magicore: Multi-agent, iterative, coarse-to-fine refinement for reasoning},
  author={Chen, Justin Chih-Yao and Prasad, Archiki and Saha, Swarnadeep and Stengel-Eskin, Elias and Bansal, Mohit},
  journal={arXiv preprint arXiv:2409.12147},
  year={2024}
}

@article{yuan2025agent,
  title={Agent-R: Training Language Model Agents to Reflect via Iterative Self-Training},
  author={Yuan, Siyu and Chen, Zehui and Xi, Zhiheng and Ye, Junjie and Du, Zhengyin and Chen, Jiecao},
  journal={arXiv preprint arXiv:2501.11425},
  year={2025}
}

@article{bahrpeyma2022review,
  title={A review of the applications of multi-agent reinforcement learning in smart factories},
  author={Bahrpeyma, Fouad and Reichelt, Dirk},
  journal={Frontiers in Robotics and AI},
  volume={9},
  pages={1027340},
  year={2022},
  publisher={Frontiers Media SA}
}

@article{tran2025multi,
  title={Multi-agent collaboration mechanisms: A survey of llms},
  author={Tran, Khanh-Tung and Dao, Dung and Nguyen, Minh-Duong and Pham, Quoc-Viet and O'Sullivan, Barry and Nguyen, Hoang D},
  journal={arXiv preprint arXiv:2501.06322},
  year={2025}
}

@article{guo2024large,
  title={Large language model based multi-agents: A survey of progress and challenges},
  author={Guo, Taicheng and Chen, Xiuying and Wang, Yaqi and Chang, Ruidi and Pei, Shichao and Chawla, Nitesh V and Wiest, Olaf and Zhang, Xiangliang},
  journal={arXiv preprint arXiv:2402.01680},
  year={2024}
}

@book{gray2011entropy,
  title={Entropy and information theory},
  author={Gray, Robert M},
  year={2011},
  publisher={Springer Science \& Business Media}
}

@aticle{li2024survey,
  title={A survey on LLM-based multi-agent systems: workflow, infrastructure, and challenges},
  author={Li, Xinyi and Wang, Sai and Zeng, Siqi and Wu, Yu and Yang, Yi},
  journal={Vicinagearth},
  volume={1},
  number={1},
  pages={9},
  year={2024},
  publisher={Springer}
}

@article{zhong2024heterogeneous,
  title={Heterogeneous-agent reinforcement learning},
  author={Zhong, Yifan and Kuba, Jakub Grudzien and Feng, Xidong and Hu, Siyi and Ji, Jiaming and Yang, Yaodong},
  journal={Journal of Machine Learning Research},
  volume={25},
  number={32},
  pages={1--67},
  year={2024}
}

@article{zeng2023mr,
  title={Mr-gsm8k: A meta-reasoning benchmark for large language model evaluation},
  author={Zeng, Zhongshen and Chen, Pengguang and Liu, Shu and Jiang, Haiyun and Jia, Jiaya},
  journal={arXiv preprint arXiv:2312.17080},
  year={2023}
}

@inproceedings{yu2025humaneval,
  title={HumanEval pro and MBPP pro: Evaluating large language models on self-invoking code generation task},
  author={Yu, Zhaojian and Zhao, Yilun and Cohan, Arman and Zhang, Xiao-Ping},
  booktitle={Findings of the Association for Computational Linguistics: ACL 2025},
  pages={13253--13279},
  year={2025}
}

@article{ralphs2003capacitated,
  title={On the capacitated vehicle routing problem},
  author={Ralphs, Ted K and Kopman, Leonid and Pulleyblank, William R and Trotter, Leslie E},
  journal={Mathematical programming},
  volume={94},
  number={2},
  pages={343--359},
  year={2003},
  publisher={Springer}
}

@article{pang2024chatgpt,
  title={ChatGPT-4o for English language teaching and learning: Features, applications, and future prospects},
  author={Pang, Samarnh and Nol, Engheang and Heng, Kimkong},
  journal={Available at SSRN 4837988},
  year={2024}
}

@article{bae2024enhancing,
  title={Enhancing software code vulnerability detection using gpt-4o and claude-3.5 sonnet: A study on prompt engineering techniques},
  author={Bae, Jaehyeon and Kwon, Seoryeong and Myeong, Seunghwan},
  journal={Electronics},
  volume={13},
  number={13},
  pages={2657},
  year={2024},
  publisher={MDPI}
}

@article{grattafiori2024llama,
  title={The llama 3 herd of models},
  author={Grattafiori, Aaron and Dubey, Abhimanyu and Jauhri, Abhinav and Pandey, Abhinav and Kadian, Abhishek and Al-Dahle, Ahmad and Letman, Aiesha and Mathur, Akhil and Schelten, Alan and Vaughan, Alex and others},
  journal={arXiv preprint arXiv:2407.21783},
  year={2024}
}

@article{hui2024qwen2,
  title={Qwen2. 5-coder technical report},
  author={Hui, Binyuan and Yang, Jian and Cui, Zeyu and Yang, Jiaxi and Liu, Dayiheng and Zhang, Lei and Liu, Tianyu and Zhang, Jiajun and Yu, Bowen and Lu, Keming and others},
  journal={arXiv preprint arXiv:2409.12186},
  year={2024}
}

@inproceedings{hong2023metagpt,
  title={MetaGPT: Meta programming for a multi-agent collaborative framework},
  author={Hong, Sirui and Zhuge, Mingchen and Chen, Jonathan and Zheng, Xiawu and Cheng, Yuheng and Wang, Jinlin and Zhang, Ceyao and Wang, Zili and Yau, Steven Ka Shing and Lin, Zijuan and others},
  booktitle={The twelfth international conference on learning representations},
  year={2023}
}

@article{wang2025unique,
  title={Unique security and privacy threats of large language models: A comprehensive survey},
  author={Wang, Shang and Zhu, Tianqing and Liu, Bo and Ding, Ming and Ye, Dayong and Zhou, Wanlei and Yu, Philip},
  journal={ACM Computing Surveys},
  volume={58},
  number={4},
  pages={1--36},
  year={2025},
  publisher={ACM New York, NY}
}

@article{he2025emerged,
  title={The emerged security and privacy of llm agent: A survey with case studies},
  author={He, Feng and Zhu, Tianqing and Ye, Dayong and Liu, Bo and Zhou, Wanlei and Yu, Philip S},
  journal={ACM Computing Surveys},
  volume={58},
  number={6},
  pages={1--36},
  year={2025},
  publisher={ACM New York, NY}
}

@article{song2025digital,
  title={Digital privacy under attack: Challenges and enablers},
  author={Song, Baobao and Pokhrel, Shiva Raj and Deng, Mengyue and Lan, Qiujun and Doss, Robin Ram and Zhu, Tianqing and Li, Gang},
  journal={ACM Computing Surveys},
  volume={58},
  number={4},
  pages={1--35},
  year={2025},
  publisher={ACM New York, NY}
}

@article{wang2025machine,
  title={When machine unlearning meets retrieval-augmented generation (rag): Keep secret or forget knowledge?},
  author={Wang, Shang and Zhu, Tianqing and Ye, Dayong and Zhou, Wanlei},
  journal={IEEE Transactions on Dependable and Secure Computing},
  year={2025},
  publisher={IEEE}
}

%\newpage

 

% \end{IEEEbiography}



\vspace{11pt}

% Optional for better spacing
\raggedbottom

\vfill

% \vfill

\end{document}